\documentclass[11pt]{article}
\usepackage[colorlinks=true,pdfborder=001,linkcolor=blue,citecolor=blue]{hyperref}
\usepackage{url}
\usepackage{stackengine}
\usepackage{nccmath}
\usepackage{mathtools}

\usepackage{mathrsfs}
\usepackage[title]{appendix}
\usepackage{adjustbox}
\usepackage{subcaption}
\usepackage{pgfplots}
\usepackage{tikz}
\usepackage{tikz,fullpage}
\usepackage{rotating}
\usepackage{tabu}
\usepackage{hhline}
\usepackage{multirow}
\usepackage{float}
\restylefloat{table}
\usepackage{adjustbox}
\usepackage[round]{natbib}
\usepackage{booktabs}
\usepackage{array}
\usepackage{tabularx}
\usepackage[utf8]{inputenc}
\usepackage{amssymb,amsmath}
\usepackage{amsfonts}
\usepackage{latexsym}
\usepackage[english]{babel}
\usepackage[T1]{fontenc}
\usepackage{eepic}
\usepackage{epic}
\usepackage{epsfig}
\usepackage{graphics,color}
\usepackage{verbatim}
\usepackage{lscape}
\usepackage{amsthm}
\usepackage{amsmath}
\usepackage{placeins}
\usepackage{float}
\usepackage{caption}
\usepackage{color}
\usepackage{setspace}
\usepackage[ruled,linesnumbered]{algorithm2e}
\usepackage{frcursive}

\DeclareMathOperator*{\argmin}{argmin}
\DeclareMathOperator*{\argmax}{argmax}
\usepackage{mathrsfs}
\usepackage{soul}
\usepackage{tikz}
\usepackage{tikz,fullpage}
\usetikzlibrary{arrows,petri,topaths,automata}
\usepackage[short,nocomma]{optidef}

\usepackage[colorinlistoftodos]{todonotes}

\newcommand{\pop}{\textsc{BIPC}}

\makeatletter

\renewcommand*\thesection{\arabic{section}}

\newif\ifbold
\boldfalse
\boldtrue
\newcommand{\bbf}{\ifbold\bgroup\bf\fi}
\newcommand{\ebf}{\ifbold\egroup\fi}

\renewcommand{\textbf}[1]{\begingroup\bfseries\mathversion{bold}#1\endgroup}
\makeatletter    
\renewcommand{\section}{\@startsection {section}{1}{\z@}%
             {-2ex \@plus -1ex \@minus -.2ex}%
             {1ex \@plus.2ex}%
             {\normalfont\Large\rmfamily\bfseries}}
\renewcommand{\subsection}{\@startsection{subsection}{2}{\z@}%
             {-1.25ex\@plus -1ex \@minus -.2ex}%
             {.75ex \@plus .2ex}%
             {\normalfont\large\rmfamily\bfseries}}
\def\@listI{\leftmargin\leftmargini       
            \parsep .25ex \@plus .1ex     
            \topsep .25ex \@plus .1ex     
            \itemsep \parsep}
\let\@listi\@listI
\@listi
\makeatother
\makeatletter

\@addtoreset{equation}{section}
\makeatother
\makeatletter

\@addtoreset{table}{section}
\makeatother

\usepackage[margin=1in]{geometry}
\usepackage{graphicx}
\graphicspath{ {./Figures/} }
\definecolor{purple}{rgb}{0.4,0.2,1}

\usepackage{titling}

\title{
\LARGE\bf Fusing Backdoors, Machine Learning, and Optimization for Large-Scale Parametric Mixed-Integer Programs
\vspace{1ex}
}

\author{\large El Mehdi Er Raqabi,$^{1,2,*}$ Pascal Van Hentenryck$^{2}$\\
\footnotesize$^1$\emph{Department of Operations and Decision Systems, Université Laval, Québec, Canada}\\
\footnotesize$^2$\emph{H. Milton Stewart School of Industrial and Systems Engineering, Georgia Institute of Technology, Atlanta, USA}\\
\footnotesize$^*$\emph{Corresponding Author: el-mehdi.er-raqabi@fsa.ulaval.ca}\\
}

\date{}

\begin{document}
\maketitle

\vspace{2cm}
\begin{abstract}
\vspace{0.5cm}

Large-scale optimization problems are often solved repeatedly under similar structural conditions, leading to substantial computational overhead. This occurs in applications such as power systems, transportation, and supply chain networks, where the underlying structure is fixed while parameters frequently vary under perturbations. 

This paper proposes a Learning to Optimize (LTO) framework that accelerates the solution of large-scale general mixed-integer problems by leveraging the concept of a backdoor, i.e., a subset of variables that drive most of the computational complexity. The proposed \pop{} framework consists of three phases. Phase I is an {\em identification} procedure that discovers a backdoor for a set of instances in the distribution. Phase II uses supervised learning to develop {\em machine learning models} that, given an instance, predict values for bounded-domain backdoor variables and intervals for wide-domain backdoor variables. These predictions define a reduced optimization problem where the predictions constrain the backdoor variables, while the other variables remain free. Phase III optimizes this reduced problem and, if necessary, applies a {\em correction step} to restore feasibility or the optimality guarantees.

Experiments on real-world, large-scale problems show substantial reductions in solution time with only a limited loss in solution quality. The framework enables organizations to solve large-scale optimization problems efficiently in the presence of frequent perturbations, such as unexpected events, demand fluctuations, or operational changes. Because these changes affect parameters rather than the problem structure, \pop{} can quickly provide high-quality, feasible solutions, offering a practical approach to integrating machine learning into existing optimization pipelines.
 
\vspace{0.2cm}
\par
\vspace{0.2cm}

\end{abstract}    

\setlength{\parindent}{1em}
\setlength{\parskip}{0.5em}
\doublespacing
\newpage

\section{Introduction}\label{section:intro}

Large-scale optimization problems play a central role in many real-world applications, including power systems, transportation, and supply chain networks. In practice, these problems are often solved repeatedly under similar structural conditions. Still, parameters such as demand, resource availability, or operational constraints change over time due to unexpected events or fluctuations. Formally, these problems are often referred to as parametric optimization problems, i.e., problems that share the same overall structure, e.g., the same generators in power systems \citep{bani2025decomposition}, the same conveyors and quays in downstream coastal planning \citep{erraqabi5}, the same facilities in package delivery \citep{mohan2025fair}, the same machines in a manufacturing layout \cite{moinuddin2026scheduling}, with only parameters changing, making numerous instances of data typically related \citep{akhlaghi2025propel,cai2025id}. These parametric problems are usually modeled using general mixed integer programs (MIPs) of the form

\[
\phi(\mathcal{P})=\argmin\{\boldsymbol{c}^\top x \mid \boldsymbol{A}x \le \boldsymbol{b},\; x \in \mathbb{R}^n,\; x_i \in \mathbb{Z}\ \forall i \in \mathcal{I}\},
\]

\noindent where $\mathcal{P}= (\boldsymbol{A}, \boldsymbol{b}, \boldsymbol{c})$ is the instance data, $\boldsymbol{A} \in \mathbb{R}^{m \times n}$, $\boldsymbol{b} \in \mathbb{R}^m$, $\boldsymbol{c} \in \mathbb{R}^n$, and $\mathcal{I} \subseteq \{1,\ldots,n\}$ denotes the index set of general integer variables. Solving each instance of this parametric problem from scratch can be computationally expensive and time-consuming, thereby limiting the ability of organizations to respond quickly to real-time operational changes, meet real-time constraints, and implement interaction optimization tools exploring many scenarios. 

This motivates the development of approaches that exploit the repetitive structure of these problems to accelerate computation while maintaining solution quality and feasibility. Large-scale optimization problems have traditionally been solved using exact methods such as MIP solvers and decomposition algorithms, or using heuristic and metaheuristic approaches \citep{thesismehdi}. To accelerate computation, practitioners often employ warm-starting, valid inequalities, and fixing techniques, as well as surrogate models that exploit problem structure \citep{wu2024towards}. More recently, learning-augmented methods have been proposed to guide optimization, for example, by predicting feasible solutions, variable bounds, or partial assignments \citep{bengio2021machine,akhlaghi2025propel}. These approaches have been applied across a range of domains, including transportation planning \citep{morabit2021machine}, supply chain network design \citep{cai2025id}, and power system operations \citep{chen2023end}, demonstrating their potential to reduce computational time while maintaining solution quality in specific settings.

Despite these advances, existing methods face limitations when applied to large-scale parametric optimization problems that are fundamentally different in nature: (1) they are expressed in terms of integer variables that can take large values, as they represent multi-faceted decisions, including procurement, production, and inventory decisions over a long time horizon; and (2) they typically feature millions of integer variables. As a result, many existing techniques that fuse operations research (OR) and machine learning (ML) are not directly applicable. Some approaches focus on predicting only integer variables taking zero values at optimality, which may not always be sufficient for real-world instances when some non-zero variables must be determined accurately to solve the problem efficiently \citep{akhlaghi2025propel,cai2025id}. The intuition in this research is that predicting the values of all integer variables $x_i \in \mathcal{I}$ is hard and costly and often yields infeasible solutions, i.e., solutions that require repairs. Instead, this paper recognizes that, for various classes of parametric MIPs, there exists a subset $\mathcal{B} \subseteq \mathcal{I}$ of integer variables, called a {\em backdoor}, that, when fixed, creates a lower-dimensional problem that is much easier to solve. To the best of the authors' knowledge, there is no general framework that systematically identifies backdoors for parametric optimization problems.

This motivates the development of the Backdoor Identification, Prediction, and Correction (\pop{}) framework for parametric MIPs that combines backdoor identification, ML, and repair mechanisms to obtain, in reasonable time, high-quality solutions to instances of parametric MIPs. This paper is a detailed presentation of \pop{} and makes the following contributions:

\begin{enumerate}
    \item \textbf{The \pop{} Framework.} The paper presents a novel 3-phase framework to obtain high-quality solutions to parametric MIPs. Phase I identifies a backdoor that drives most of the computational complexity for a class of instances. Phase II uses ML to predict values or restricted domains for the backdoor variables. Phase III uses optimization to complete the partial assignment and possibly corrects the predictions using repair mechanisms.  
    
    \item \textbf{The \pop{} Predictions.} Unlike existing methods for MIP models that often predict only zeros for integer variables or focus on 0-1 MIPs, \pop{} predicts all backdoor variables. For integer variables spanning a short range, \pop{} predicts a specific value. For integer variables spanning a large range, it predicts 0 or restricted domains. \pop{} thus generalizes existing methods in the literature \citep{akhlaghi2025propel,cai2025id}.
    
    \item \textbf{The \pop{} Taxonomy.} The paper presents a taxonomy for designing \pop{} for a given parametric optimization MIP. The taxonomy includes the problem size, variable types, and feature types. 
    
    \item \textbf{\pop{} Applications and Computational Results.} \pop{} is tested on five real-world large-scale problems: a middle-mile consolidation network design (MMCNP), a strategic locomotive assignment problem (SLAP), exam class scheduling (COURSE), a downstream supply chain planning (OCP), and an order fulfillment problem (OFP). The results demonstrate consistent and significant reductions in solution time while maintaining solution quality.
\end{enumerate}

The remainder of the paper is structured as follows. Section~\ref{sec:literature} reviews the relevant literature on learning-based methods and proxy approaches. Section~\ref{sec:framework} presents an overview of \pop{}. Section~\ref{sec:taxonomy} introduces the \pop{} taxonomy and methodology. Sections~\ref{sec:proxy_phase}--\ref{sec:correction_phase} discuss the three \pop{} phases in detail. Section~\ref{sec:experiments} describes the experimental design and the real-world large-scale problems considered. Section~\ref{sec:results} reports computational results and analyzes the performance of the \pop{} framework. Finally, Section~\ref{sec:conclusion} concludes the paper and outlines directions for future research.

\section{Literature Review}\label{sec:literature}

The use of ML for MIPs has attracted significant attention in recent years. It is increasingly recognized as a promising direction to overcome some of the computational challenges \citep{bengio2021machine,kotary2021end,vesselinova2020learning}, especially for problems that can only be solved by fusing the powers of OR and ML \citep{van2024ai4opt}. Learning-based approaches are well known for their ability to yield effective empirical algorithms that leverage regularities in large real-world datasets and improve solution times \citep{cappart2021combining,kotary2021end,morabit2021machine,khalil2022mip}. 

The integration of optimization and ML recently led to the development of several distinct approaches, surveyed by \cite{kotary2021end}. These methods, often referred to by various names in the literature, can be categorized into three main categories, increasingly hybridized: 1) Decision-focused learning, 2) Optimization proxies, and 3) Learning to optimize. Decision-focused learning, also called smart-predict-then-optimize or predict-and-search, aims at training forecasting and optimization models in the same pipeline (see e.g., \cite{elmachtoub2022smart,el2019generalization,ning2019optimization,sahinidis2004optimization,donti2017task}). Optimization proxies aims at learning the mapping between the input and an optimal solution of an optimization problem (see e.g., \cite{chen2023end,kotary2021end,kotary2022fast,donti2021dc3,huang2021deepopf,park2023self}). Learning to optimize focuses on techniques to improve the efficiency and the solution quality of optimization algorithms and solvers \citep{tang2024learn,maragno2025mixed}. This paper is a hybridization of these three categories and is the first to leverage their combined capabilities.

\noindent \textbf{Decision-focused learning.} In decision-focused learning, \cite{park2023confidence} introduce the Predict-Repair-Optimize framework, which predicts an optimal solution, fixes a subset of variables with confident predictions, and restores feasibility with a dedicated repair algorithm. A final optimization step is applied to complete the partial assignment. \cite{han2023gnn} propose a novel predict-and-search (PaS) framework that adopts the trust region method, which searches for near-optimal solutions within a well-defined region. Their approach uses a trained graph neural network (GNN) to predict marginal probabilities of binary variables in a given MIP instance. It then searches for near-optimal solutions within the trust region using these predictions. \cite{huang2024contrastive} propose the ConPaS framework that learns to predict solutions to MIPs with contrastive learning. ConPaS collects both high-quality solutions as positive samples and low-quality or infeasible solutions as negative samples, and learns to make discriminative predictions by contrasting these samples. It then fixes the assignments for a subset of integer variables and solves the resulting reduced MIP to obtain high-quality solutions. ConPaS was evaluated on four classes of binary MIPs. Most of these approaches primarily evaluate their algorithms on binary problems. \cite{cai2025id} extend the PaS framework to parametric MIPs and introduce \textsc{ID-PaS+}, an identity-aware learning framework that enables the ML model to handle heterogeneous variables more effectively. Experiments on several real-world large-scale problems demonstrate that \textsc{ID-PaS+} consistently outperforms the state-of-the-art solvers Gurobi and PaS.

\noindent \textbf{Optimization Proxies.} Optimization proxies have accumulated great success for continuous optimization problems. \cite{tanneau2024dual} present dual Lagrangian learning (DLL), a principled learning methodology for dual conic optimization proxies. The proposed methodology significantly outperforms a state-of-the-art learning-based method and achieves 1000$\times$ speedups over commercial interior-point solvers with optimality gaps under 0.5\% on average. However, they encounter feasibility and training challenges when discrete variables are present \citep{tran2021differentially,fioretto2020predicting,detassis2021teaching,guan2026proxy}. Learning models for discrete optimization problems lack useful gradients because the arg-max operator is piecewise constant, which complicates backpropagation. One way to address this is to construct effective approximations of the gradients, e.g., by generating continuous surrogates of the MIP to facilitate effective training \citep{ferber2020mipaal,kotary2021end,donti2017task,wilder2019melding}. This paper combines a partial proxy with a correction phase to handle the discrete context. The focus is on the backdoor variables $x_i, \ i \in \mathcal{B}$, which drive complexity. Then, the solution is completed. The correction phase restores feasibility in case the obtained solution is infeasible.

\noindent \textbf{Learning to Optimize.} Learning to optimize encompasses a wealth of approaches, many of which are reviewed in \cite{lodi2017learning}, \cite{bengio2021machine}, and \cite{kotary2021end}. They include techniques to guide search decisions in branch-and-bound/cut solvers \citep{zarpellon2021parameterizing,gasse2019exact,gupta2020hybrid,tang2020reinforcement} and direct the application of primal heuristics within branch-and-bound (\cite{khalil2017learning,chmiela2021learning, bengio2020learning,song2020general,ye2026democratizing}). \cite{khalil2016learning} highlight that, beyond the supervised learning approaches prevalent in this context, reinforcement learning formulations are worth exploring due to the sequential nature of the variable selection task. Aligned with this observation, \cite{akhlaghi2025propel} implement a novel approach that combines the benefits of supervised learning and deep reinforcement learning. They propose \textsc{PROPEL}, a new framework that integrates optimization with both supervised and deep reinforcement learning (DRL) to reduce the search space significantly. Although it differs from these studies, \textsc{PROPEL} shares some similarities with those focusing on variable selection strategies (\cite{khalil2016learning,alvarez2017machine,pmlr-v80-balcan18a}). \textsc{PROPEL} uses supervised learning not to predict the values of all integer variables, but to identify the variables fixed to zero in the optimal solution. \textsc{PROPEL} drew inspiration from \cite{park2023confidence}, but differs in two key aspects. First, it fixes only the variables to zero, leaving the nonzero integer variables free. Second, it uses deep reinforcement learning to determine which variables to unfix to obtain the desired optimality target.

\noindent \textbf{Paper Positioning.}
This work lies at the intersection of learning-based MIP acceleration and the backdoor literature in combinatorial optimization. The first stream learns solution information or solver decisions for MIPs. The \textsc{PROPEL} framework is the closest to this research: 1) it is the first framework that fuses OR and ML for industrial supply chain planning, 2) it considers problems with arbitrary integers, and 3) it is not restricted to graph-based problems and applies to standard MIP formulations. However, it focuses only on predicting variables that take the value 0 among $x_i \in \mathcal{I}$ and does not consider the concept of {\em backdoor}. Furthermore, \textsc{PROPEL} has been evaluated in a single case study of industrial supply chain planning. The authors highlight the need to consider other classes of applications beyond supply chain planning and other solution techniques for both stages in the \textsc{PROPEL} framework, and the potential for even larger problems than the supply chain planning instances \citep{akhlaghi2025propel}.

The second stream studies backdoors as structural explanations for solver performance. Backdoors were originally introduced in the Boolean satisfiability problem (SAT) and the constraint satisfaction problem (CSP) as sets of variables whose instantiation makes the remaining problem tractable \citep{gomes2000heavy,williams2003backdoors}. This idea was later extended to optimization, where different notions of feasibility and optimality backdoors were proposed for combinatorial optimization and mixed-integer programming \citep{dilkina2007tradeoffs,dilkina2009backdoors}. More recent work learns effective MILP backdoors using ranking or contrastive learning, but these methods use the predicted backdoor to guide branching decisions inside branch-and-bound rather than to predict variable assignments or domains \citep{khalil2022finding,cai2024learning}.

\pop{} differs from both streams. It identifies a stable computational backdoor at the level of a parametric MIP family, learns values or intervals for the backdoor variables, and uses these predictions to construct a reduced optimization problem that is then completed and corrected. Thus, the backdoor in \pop{} is not merely a branching guide. It is the interface between learning and optimization.

\pop{} extends {\textsc PROPEL} along six axes: 1) \textbf{Backdoor Variables.} \pop{} is organized around the concept of backdoor: it identifies a backdoor for a parametric MIP and organizes the prediction phase around this backdoor. 2) \textbf{Prediction.} \pop{} predicts not only zero-valued variables; it also predicts the values or restricted domains for non-zero variables, 3) \textbf{Real Case Studies.} \pop{} is evaluated with five different large-scale real-world applications, 4) \textbf{Scalability.} This paper tackles problems with millions of variables and constraints, addressing problem sizes that are significantly larger than those considered in the existing literature, 5) \textbf{Transferability.} This research broadens the class of MIPs and applications that can benefit the fusion of OR and ML to five applications, and 6) \textbf{Feasible and High-Quality Solutions.} This research reduces the size of the search space by focusing on the backdoor variables and uses correction mechanisms to decide which variables to relax, potentially enhancing feasibility and near-optimality.

To the best of the authors' knowledge, \pop{} is the first framework to combine backdoor identification, value or interval prediction, and optimization-based completion for large-scale parametric MIPs with general integer variables.

\section{Overview of the \pop{} Framework}\label{sec:framework}

Consider the parametric mixed-integer optimization problem
\[
\phi(\mathcal{P})=\argmin\{\boldsymbol{c}^\top x \mid \boldsymbol{A} x \le \boldsymbol{b},\; x \in \mathbb{R}^n,\; x_i \in \mathbb{Z}\ \forall i \in \mathcal{I}\},
\]
where multiple instances share the same structural formulation but differ in the instance data $\mathcal{P} = (\boldsymbol{A}, \boldsymbol{b}, \boldsymbol{c})$ with $\boldsymbol{A} \in \mathbb{R}^{m \times n}$, $\boldsymbol{b} \in \mathbb{R}^m$, $\boldsymbol{c} \in \mathbb{R}^n$, and $\mathcal{I} \subseteq \{1,\ldots,n\}$ denotes the index set of general integer variables. The instance data is taken from a distribution ${\cal D}$, learned from historical data and/or forecasts.

The \pop{} framework assumes that a subset $\mathcal{B} \subseteq \mathcal{I}$ of variables drives most of the complexity in solving instances of $\phi$. These variables are often called a {\em backdoor} in the artificial intelligence (AI) literature, and this article adopts this terminology. Informally speaking, \pop{} first identifies this backdoor using a reference instance and then develops ML models to predict their values. \pop{} then solves an optimization problem to complete this partial solution. 

\begin{figure}[t!]
    \centering
    \includegraphics[width=\linewidth]{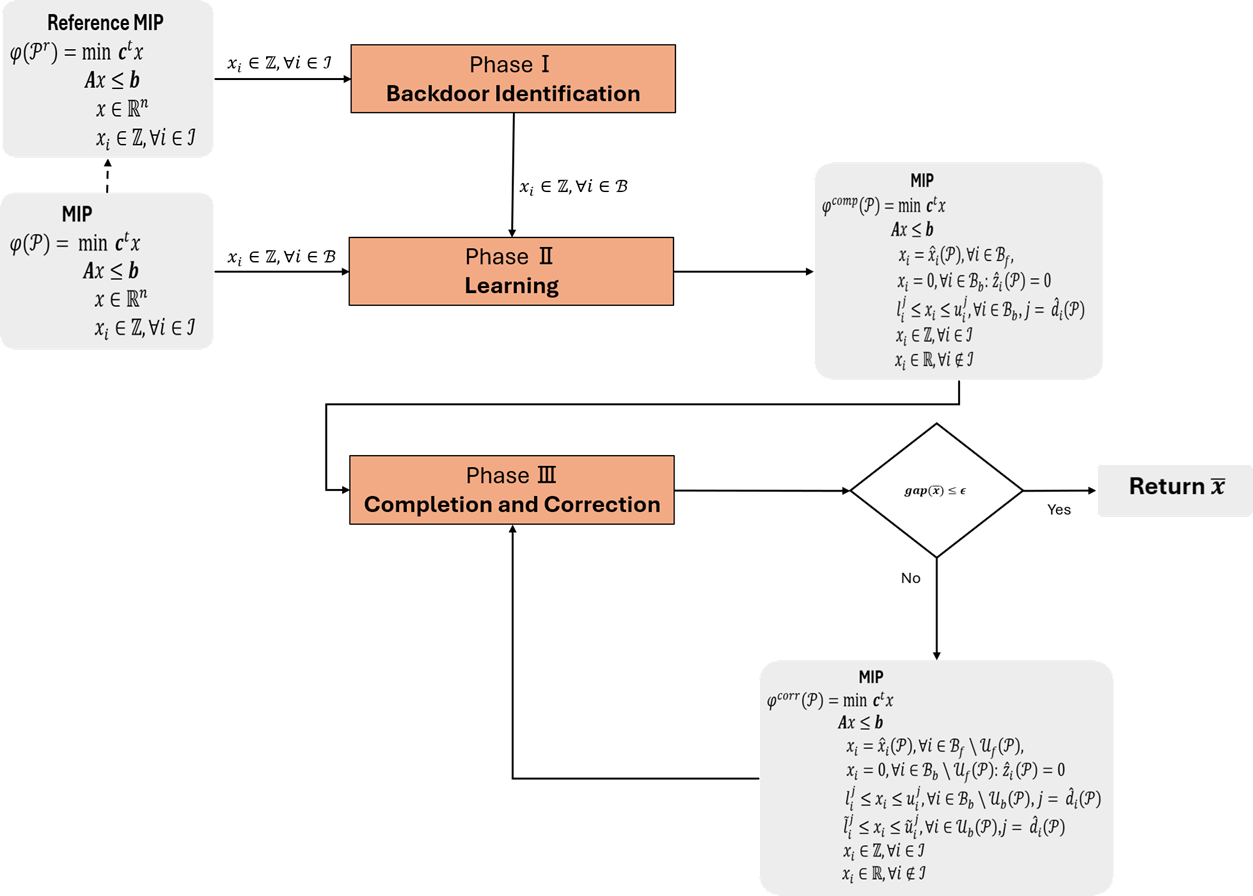}
    \caption{The \pop{} Framework.}
    \label{fig:pop_framework}
\end{figure}

More precisely, \pop{} consists of three phases: (I) backdoor identification, (II) prediction, and (III) completion and correction. The identification phase tries to discover the backdoor ${\cal B}$, which is partitioned into two sets: 
${\cal B}_f$ and ${\cal B}_b$. The prediction phase builds ML models that predict, for an instance $\mathcal{P}$, the values of the variables from ${\cal B}_f$ in the optimal solution $\phi(\mathcal{P})$ and intervals
for the variables from ${\cal B}_b$ which contain the optimal solution. The completion and correction phase uses the ML models and solves a reduced optimization problem where the variables in ${\cal B}_f$ are fixed to their predicted values and the variables in ${\cal B}_b$ are constrained to be within their predicted intervals. This reduced problem can be infeasible and suboptimal, in which this phase applies corrections to the predictions until the problem becomes feasible and satisfies the optimality tolerance. Importantly, phases I and II are performed offline using the distribution ${\cal D}$ (training time), while phase III is executed online for new instances (inference time).  The rest of this section goes into more details of each of these phases. A visual description of \pop{} is given in Figure \ref{fig:pop_framework}. 

\paragraph*{Backdoor Identification.} 

The identification phase aims at finding the backdoor ${\cal B}$. \pop{} assumes the availability of a reference instance ${\cal P}^r$ and that the backdoor remains essentially the same for all instances of distribution ${\cal D}$. This assumption is motivated by many practical optimization settings in which the same class of strategic decisions repeatedly drives computational complexity, even as the instance data changes over time. For example, in facility-location problems, facility-opening variables often determine the overall structure of the solution. In production planning, production quantities across products and periods typically drive most operational decisions. In network design problems, arc activation or flow variables frequently dominate the combinatorial complexity. As a result, while parameter values may vary from one instance to another, the variable family acting as the backdoor often remains unchanged. Under this assumption, the backdoor identified from the reference instance can be used throughout the learning and optimization phases.

The identification phase first partitions the set of variables $\mathcal{I}$ into $\{\mathcal{I}_1,\ldots,\mathcal{I}_m\}$ based on the structure of $\phi$. It then solves the reference instance $\phi({\cal P}^r)$ to obtain a reference solution $x^r$. The goal of the identification phase is to determine which subset $\mathcal{I}_k$ is a potential backdoor. To identify this subset, \pop{} solves $m$ reduced optimization problems, one for each subset. The optimization problem associated with subset $\mathcal{I}_k$ $(1 \leq k \leq m)$ is the instance $\phi({\cal P}^r)$ where each variable $x_j$ in $\mathcal{I}_k$ is fixed to its value in the reference solution $x^r$, i.e., $x_j = x^r_j$. Since the objective of \pop{} is to accelerate optimization, the subset for which an optimal solution is found the fastest is considered the backdoor. The pseudo-code of the backdoor identification is given in Algorithm \ref{algo:backdoor}. It uses a function $\phi({\cal P},S)$ which returns a pair $(x^*,t^*)$ where $x^*$ is an optimal solution to instance ${\cal P}$ with additional constraints $S$ and $t^*$ is the computation time of the optimization.

\SetKwInput{KwInput}{Input}
\SetKwInput{KwOutput}{Output}

\begin{algorithm}[t!]
\KwInput{$\phi$, ${\cal P}^r$, $\{\mathcal{I}_1,...,\mathcal{I}_m\}$}
\SetAlgoLined
$t^* = \infty$ \\
$x^r$ = $\phi({\cal P}^r)$ \\
\For{$i \in  \{1,...,m\}$}{
    $S = \{ x_k = x_k^r \ | \ k \in {\cal I}_i \}$ \\
    $(x^i,t^i)$ = $\phi({\cal P}^r,S)$\\
    \If{$t^i < t^*$}{
        $t^* = t^i$ \\
        $i^* = i$
    }
}
\Return{$\mathcal{I}_{i^*}$}  
\caption{Phase I: The Backdoor Identification Procedure.}
\label{algo:backdoor}
\end{algorithm}

\paragraph*{The Learning Phase.} 
Given a backdoor ${\cal B}$, the learning phase starts by partitioning its variables into ${\cal B}_f$ and  
${\cal B}_b$. It then learns two ML models ${\cal M}_f$ and ${\cal M}_b$ for ${\cal B}_f$ and  
${\cal B}_b$ respectively. ${\cal M}_f$ learns the mapping $\phi$ but only for the variables in  ${\cal B}_f$.
${\cal M}_b$ learns to predict intervals for the variables in ${\cal B}_b$ which contain the optimal solutions for a
given instance. ${\cal M}_b$ is a classification problem where the domains of the variables in ${\cal B}_b$ have been partitioned into a set of intervals. 

\paragraph*{Completion and Correction.} 
At inference time, given a new instance ${\cal P}$, \pop{} uses ${\cal M}_f$ and ${\cal M}_b$ to predict the values and intervals for ${\cal B}_f$ and ${\cal B}_b$ in the optimal solution $\phi({\cal P})$. For instance $\mathcal{P}$, \pop{} then solves the reduced optimization problem 
\[
\phi({\cal P},\{ x_k = \hat{x}_k(\mathcal{P}) \ | \ k \in {\cal B}_f \} \ \cup \ \{ x_k \in \hat{D}_k(\mathcal{P}) | \ k \in {\cal B}_b \})
\]
which fixes each variable $x_k$ in  ${\cal B}_f$ to its predicted value $\hat{x}_k(\mathcal{P})$ and constrains each variable $x_k$ in ${\cal B}_b$ to be in its predicted intervals $\hat{D}_k(\mathcal{P})$. This reduced problem may be infeasible or suboptimal, in which case a correction phase is applied. The correction selectively relaxes a subset of variables in $\mathcal{B}$, restoring feasibility while preserving as much of the predictions as possible. Variables in $\mathcal{B}_{f}$ are relaxed by removing their assigned values. Variables in $\mathcal{B}_{b}$ are relaxed by enlarging their intervals.

Since the design of the ML models and correction mechanisms is problem-dependent, the next section focuses on the taxonomy and methodology used in the \pop{} framework to specify the modeling choices, feature design, and algorithmic strategies used for the three phases across different problem classes.

\section{Taxonomy and Methodology}\label{sec:taxonomy}

Before detailing the methodology for predicting and correcting the backdoor, this section introduces a conceptual taxonomy that organizes model and strategy types. The taxonomy highlights the key factors underlying the design space of Phases II and III, providing a systematic basis for selecting appropriate ML models and correction mechanisms within the \pop{} framework. The section presents the taxonomy first before outlining the methodology for designing the ML models and correction mechanisms.

\subsection{Taxonomy of Models and Strategies}\label{sec:taxonomy_dimensions}

\begin{figure}[!t]
\centering
\begin{tikzpicture}
\begin{axis}[
    xlabel={Variable Type},
    ylabel={Problem Size},
    zlabel={Feature Type},
    xtick={1,2},
    xticklabels={$V_1$, $V_2$},
    ytick={1,2},
    yticklabels={$P_1$, $P_2$},
    ztick={1,2},
    zticklabels={$F_1$, $F_2$},
    zlabel style={sloped},
    grid=major,
    view={135}{25},
    zticklabel style={xshift=-1pt},  
    tick label style={font=\small},
    axis lines=left,
    grid=both,
]
\addplot3[
    only marks,
    mark=cube*,
    mark size=6,
]
coordinates {
    (1,1,1) (2,1,1)
    (1,2,1) (2,2,1)

    (1,1,2) (2,1,2)
    (1,2,2) (2,2,2)
};

\addplot3[
    only marks,
    mark=cube*,
    mark size=6,
    fill=blue!65,
    draw=black,
]
coordinates {
    (1,2,1)
};

\addplot3[
    only marks,
    mark=cube*,
    mark size=6,
    fill=red!65,
    draw=black,
]
coordinates {
    (2,2,1) 
};

\addplot3[
    only marks,
    mark=cube*,
    mark size=6,
    fill=green!65,
    draw=black,
]
coordinates {
    (1,2,2) 
};

\addplot3[
    only marks,
    mark=cube*,
    mark size=6,
    fill=purple!65,
    draw=black,
]
coordinates {
    (2,2,2) 
};

\addplot3[
    only marks,
    mark=cube*,
    mark size=6,
    fill=orange!70,
    draw=black,
]
coordinates {
    (1,1,1) 
};

\addplot3[
    only marks,
    mark=cube*,
    mark size=6,
    fill=yellow!70,
    draw=black,
]
coordinates {
    (2,1,1) 
};

\addplot3[
    only marks,
    mark=cube*,
    mark size=6,
    fill=gray!70,
    draw=black,
]
coordinates {
    (1,1,2)
};

\addplot3[
    only marks,
    mark=cube*,
    mark size=6,
    fill=pink!70,
    draw=black,
]
coordinates {
    (2,1,2)
};
\end{axis}
\end{tikzpicture}
\caption{Visualization of the \pop{} framework taxonomy across three dimensions: variable type ($V_1$, $V_2$), problem size ($P_1$, $P_2$), and feature type ($F_1$, $F_2$). Each colored cube corresponds to a distinct configuration of the taxonomy.}
\label{fig:pop_taxonomy}
\end{figure}
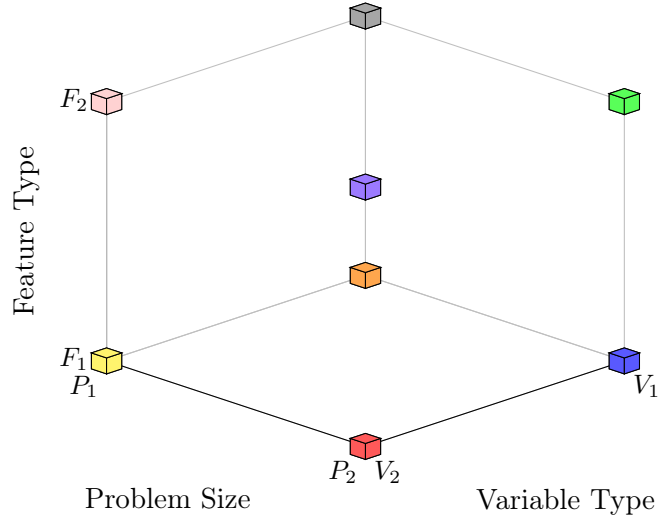

To systematically address the diversity of problem instances and guide the design of Phases II and III, the \pop{} taxonomy classifies instances along three dimensions: \emph{problem size}, \emph{variable type}, and \emph{feature type} (see Figure \ref{fig:pop_taxonomy}).

\noindent \textbf{Problem Size.}
This captures the scale of the problems which are divided into two classes:
\begin{itemize}
\item \textit{Medium-Scale ($P_1$)}: Problems with a moderate number of variables, i.e., problems that can be handled effectively by a single predictive model.
\item \textit{Large-Scale ($P_2$)}: Problems with a very large number of variables, i.e., problems whose scale motivates variable-wise prediction to improve scalability.
\end{itemize}

\noindent \textbf{Variable Type.}  
This captures the nature of backdoor variables, which are divided into two classes:
\begin{itemize}
    \item \textit{Bounded-Range ($V_1$)}:  binary and integer variables with a limited domain.
    \item \textit{Wide-Range ($V_2$)}: Integer variables with large domains.
\end{itemize}

\noindent \textbf{Feature Type.}
This captures the choice of input features, which are divided into two classes:
\begin{itemize}
\item \textit{Without Domain-Specific Features ($F_1$)}: Models only rely on generic instance data, such as the constraint matrix, objective coefficients, relaxation objective, etc.
\item \textit{With Domain-Specific Features ($F_2$)}: Models augment generic instance features with additional features informed by domain knowledge.
\end{itemize}

\noindent
This taxonomy provides a structured framework for mapping problem characteristics to ML models and correction strategies, allowing for systematic design choices in Phases II and III. The next subsection discusses the methodology for constructing, training, and applying the ML models, as well as the associated correction mechanisms, based on this taxonomy.

\subsection{Methodology for Phases II and III}
\label{sec:methodology}

Building on the taxonomy above, the methodology for constructing and applying the ML model (Phase II) and the correction mechanism (Phase III) is guided by problem size, variable type, and feature type. The following summary outlines the general approach, highlighting how these three dimensions influence the choice of the strategies. The choices are summarized in Table \ref{tab:pop_cube_table}.

\begin{table}[t!]
\centering
\caption{Mapping of colored cubes in Figure~\ref{fig:pop_taxonomy} to Phases II and III settings. Phase II includes ML model architecture, type, and features. Phase III indicates the correction strategy.}
\label{tab:pop_cube_table}
\begin{tabularx}{0.8\textwidth}{ccccccc}
\hline
\textbf{ID} & \textbf{Color} & \textbf{Code} & \textbf{Architecture} & \textbf{Type} & \textbf{Feature} & \textbf{Correction} \\
\hline
1 & Orange & $P1V1F1$ & $G$ & $VP$ & $GF$ & $SR$ \\
2 & Gray & $P1V1F2$ & $G$ & $VP$ & $CF$ & $SR$\\
3 & Yellow & $P1V2F1$ & $G$ & $VP\&IP$ & $GF$ & $SR \& BA$\\
4 & Pink & $P1V2F2$ & $G$ & $VP\&IP$ & $CF$ & $SR \& BA$\\
5 & Blue & $P2V1F1$ & $VW$ & $VP$ & $GF$ & $SR$\\
6 & Green & $P2V1F2$ & $VW$ & $VP$ & $CF$ & $SR$\\
7 & Red & $P2V2F1$ & $VW$ & $VP\&IP$ & $GF$ & $SR \& BA$\\
8 & Purple & $P2V2F2$ & $VW$ & $VP\&IP$ & $CF$ & $SR \& BA$\\
\hline
\end{tabularx}
\end{table}

\noindent \textbf{ML Architecture.} The design of the ML models depends on the problem size:
\begin{itemize}
\item \textit{Global Model ($G$)}: For medium-scale problems ($P_1$), a single model predicts all variables simultaneously, ensuring consistency across variable interactions.
\item \textit{Variable-wise Model ($VW$)}: For large-scale problems ($P_2$), a separate model per variable captures individual variable behaviors, allowing scalability and specialized architectures at the cost of losing variable interactions.
\end{itemize}

\noindent \textbf{Model Type.} The treatment of backdoor variables depends on whether they are small or large domains:
\begin{itemize}
    \item \textit{Value Prediction ($VP$)}: For problems of type $V_1$, predicting values is the selected method. Predicted values are fixed in the reduced problem, and Phase III handles any resulting feasibility issues.
    \item \textit{Interval Prediction ($IP$): For problems of type $V_2$, in addition to classifying whether a variable is zero versus non-zero, the ML model predicts an interval for the value of each variable when it is non-zero. Rather than estimating exact values, the method learns data-driven bounds that restrict the domain of non-zero variables in the reduced problem. Phase III then selectively relaxes these intervals, if needed, to restore feasibility and maintain solution quality.}
\end{itemize}

\noindent \textbf{ML Model Features.} Input features influence model accuracy and robustness:
\begin{itemize}
    \item \textit{Generic-Feature Model ($GF$)}: Models use only standard instance data, such as constraint matrices or objective coefficients.
    \item \textit{Customized-Feature Model ($CF$)}: Models combine generic features with additional problem-specific information, such as structural properties, aggregated metrics, or historical solution patterns.
\end{itemize}

\noindent \textbf{Correction Mechanism.} Phase III ensures robustness regardless of ML model design using the following strategies:
\begin{itemize}
    \item \textit{Selective Relaxation ($SR$)}: Only the subset of predicted variables causing infeasibility or large optimality gaps is relaxed, preserving the learning model structure as much as possible.
    \item \textit{Bound Adjustment ($BA$)}: For wide-range integer variables, the predicted intervals may be progressively adjusted by expanding their lower and upper bounds, e.g., by incorporating adjacent intervals, to restore feasibility while keeping predictions informative.
\end{itemize}

Phase III of \pop{} focuses only on corrections that act on variable values and their bounds, as these are the only objects predicted in Phase II. This design ensures methodological alignment and preserves a clear interpretation of the ML model as defining a reduced optimization problem.

Many real-world large-scale optimization problems are sparse, with most backdoors equal to zero at optimality. In such cases, the primary challenge is correctly identifying the non-zero variables, as wrong predictions on zero-valued variables typically have limited impact. Accordingly, Phase III corrections focus on wrongly predicted non-zeros, i.e., either by relaxing incorrectly fixed variables or adjusting their bounds, while largely preserving the ML structure for zero-valued variables.

The proposed methodology enables \pop{} to efficiently predict and correct backdoor variables across diverse parametric mixed-integer optimization problems. The taxonomy in Section~\ref{sec:taxonomy_dimensions} guides these design decisions and maps them to the colored cubes in Figure~\ref{fig:pop_taxonomy}, providing a clear reference for Phases II and III.

\section{The ML Models}
\label{sec:proxy_phase}

This section discusses the learning task, supervised learning, and the reduced MIP problem. The ML models use a set of labeled instances $\{(\mathcal{P}^k,\phi(\mathcal{P}^k))\}_{k=1}^K$ drawn from distribution $\mathcal{D}$.

\subsection{Variables with Bounded Domains}
\label{sec:proxy_learning}

\pop{} uses supervised learning with a cross-entropy loss to predict the values of variables in ${\cal B}_f$. Given a domain $D_i$ for variable $x_i$ and an instance ${\mathcal P}$, model ${\cal M}_f$ returns a (softmax) probability distribution ${\cal M}_f(\mathcal{P},x_i)$ over $D_i$. The loss function for variable $x_i$ and instance ${\cal P}$ is given by 
\[
\mathcal{L}^f_i(\mathcal{P}) =
-\sum_{v \in D_i} \mathbf{1}\{\phi(\mathcal{P})_i = v\} \;
\log ({\cal M}_f(\mathcal{P},x_i)_v)
\]

\subsection{Variables with Large Domains}

For variables in ${\cal B}_b$ with large domains, \pop{} uses two ML models: ${\cal M}_b^0$ and ${\cal M}_b^I$. Model ${\cal M}_b^0$ is a binary classifier that determines whether a variable in ${\cal B}_b$ is zero. Model ${\cal M}_b^0$ also uses a cross-entropy loss for each variable $x_i$, i.e., 
\[
\mathcal{L}^b_i(\mathcal{P}) =
-\sum_{v \in \{0,1\}} \mathbf{1}\{\psi(\mathcal{P})_i = v\} \;
\log ({\cal M}_b^0(\mathcal{P},x_i)_v)
\]
where 
\[
\psi(\mathcal{P})_i :=
\begin{cases}
0, & \text{if } \phi(\mathcal{P})_i = 0, \\
1, & \text{otherwise}.
\end{cases}
\]
Model ${\cal M}_b^I$ predicts an interval (or domain) for each of the variables in ${\cal B}_b$ that are not predicted to be zero in the optimal solution of an instance. \pop{} partitions the domain $D_i$ of each variable $x_i \in {\cal B}_b$ into a set of intervals, i.e., 
$
I_i = \{[l_i^1, u_i^1], \ldots, [l_i^p, u_i^p]\}.
$
Given a domain $D_i$ for variable $x_i$ and an instance ${\cal P}$, model ${\cal M}_b^I$ returns a (softmax) probability distribution  over $I_i$,
where ${\cal M}_b^I({\cal P}, x_i)_j$ denotes the probability that optimal solution $\phi({\cal P})_i$ lies in $[l_i^j,u_i^j]$. Again, model ${\cal M}_b^I$ is trained with a cross-entropy loss, i.e.,
{\[
\mathcal L_i^I(\mathcal P)
=
-\sum_{j=1}^{p}
\mathbf 1\{\phi(\mathcal P)_i \in [l_i^j,u_i^j]\}
\log({\cal M}_b^I(\mathcal P,x_i)_j)
\]}

\subsection{Architecture} The deep neural net (DNN) architectures used for both classification and interval prediction are designed with problem size in mind. For medium-scale problems ($P_1$) with a moderate number of backdoor variables, a single global model predicts all variables simultaneously, capturing dependencies and interactions across variables, while remaining computationally efficient. In contrast, for large-scale problems ($P_2$) with many backdoor variables, a separate model is trained for each variable. This approach enables scalability and the use of specialized architectures or hyperparameters for each variable, which is particularly useful when variables differ in sparsity patterns or range.

In addition to problem size, the choice of input features influences model design. All DNNs can be trained using only generic instance-level data, such as the constraint matrix $\boldsymbol{A}$ and objective coefficients $\boldsymbol{c}$. When available, domain-specific features can be incorporated to improve predictive accuracy, reducing dependencies on Phase III corrections. This design ensures that the architecture is flexible and consistent across value and interval prediction tasks in Phase II.

These predictions, whether obtained from global or variable-wise DNNs and based on generic or customized features, define the reduced optimization problem solved in Phase III and therefore constitute the interface between learning and optimization in the \pop{} framework.

\section{The Reduced MIP Model and the Completion Phase}

 At inference time, \pop{} predicts the value $\hat{x}_i(\mathcal{P})$ of variable $x_i$ in ${\mathcal B}_f$ for an instance ${\cal P}$ by selecting the most likely value, i.e.,
\[
\hat{x}_i(\mathcal{P}) \in \argmax_{v \in D_i} \ {\cal M}_f(\mathcal{P},x_i)_v.
\]
\pop{} fixes these variables to their predicted value in the reduced optimization problem.
For variable $x_i$ in ${\mathcal B}_b$ and instance ${\cal P}$, \pop{} first predicts whether the variable is zero in the optimal solution of ${\cal P}$, i.e.,
\[
\hat{z}_i(\mathcal{P}) \in \argmax_{v \in \{0,1\}} \ {\cal M}_b^0(\mathcal{P},x_i)_v.
\]
If $\hat{z}_i(\mathcal{P}) = 0$, variable $x_i$ is fixed to zero in the reduced optimization problem. Otherwise, \pop{} selects the most likely interval for variable $x_i$, i.e., 
\[
\widehat{d}_i(\mathcal{P}) \in \argmax_{j \in \{1,\ldots,p\}} \ {\cal M}_b^I(\mathcal{P},x_i)_j.
\]
Variable $x_i$ is constrained to be in $[l_i^j,u_i^j]$, with $j = \widehat{d}_i(\mathcal{P})$ in the reduced optimization problem.

The initial reduced optimization problem of \pop{} is depicted in Figure \ref{fig:reduced-Optimization}: it uses the predictions listed above to constrain the backdoor variables. All other variables remain free and constrained within their initial domains. Solving $\phi^r(\mathcal{P})$ yields a completed solution $\bar{x}$, which may subsequently be corrected in case of infeasibility or significant sub-optimality. The reduced optimization problem is the core mechanism through which \pop{} combines learning and optimization: the ML models restrict the search space through backdoor predictions, while optimization completes the remaining decisions and preserves feasibility.

\begin{figure}[!t]
\allowdisplaybreaks
\begin{align}
\phi^r(\mathcal{P}) = 
\argmin_{x} \;& \boldsymbol{c}^\top x \nonumber \\
\text{s.t. } & \boldsymbol{A} x \le \boldsymbol{b}, \nonumber \\
& x_i = \hat{x}_i(\mathcal{P}), && \forall i \in \mathcal{B}_f, \nonumber \\ 
& x_i = 0 && \forall i \in \mathcal{B}_b: \hat{z}_i(\mathcal{P}) = 0 \nonumber \\ 
& l_i^j \le x_i \le u_i^j, && \forall i \in \mathcal{B}_b, j = \widehat{d}_i(\mathcal{P}) \nonumber \\
& x_i \in \mathbb{Z}, && \forall i \in \mathcal{I}, \nonumber \\
& x_i \in \mathbb{R}, && \forall i \notin \mathcal{I}, \nonumber
\end{align}    
\caption{The Core Reduced Optimization Problem of \pop{}.}
\label{fig:reduced-Optimization}
\end{figure}
\noindent

\section{The Correction Phase}\label{sec:correction_phase}

\noindent Phase II typically yields reduced MIP problems whose solutions are feasible and near-optimal \citep{akhlaghi2025propel}. Nevertheless, in some instances, fixing or bounding the backdoor variables can produce infeasible solutions or larger-than-desired optimality gaps. To address this, Phase III introduces a correction mechanism that selectively relaxes or adjusts variable bounds. This ensures feasibility and improves solution quality, while preserving the structural guidance provided by Phase II. This section discusses how to relax subsets of fixed variables 
and adjust their intervals.

\subsection{Selective Relaxation}\label{sec:selective_relaxation}

Selective relaxation targets a subset of variables initially fixed by the ML model, but they are likely causing infeasibility or suboptimality. Define $\mathcal{U}_{f}(\mathcal{P}) \subseteq \mathcal{B}$ as the subset selected for relaxation. The selectively relaxed problem is then

\begin{align}
\phi^{sr}(\mathcal{P}) = 
\argmin_{x} \;& \boldsymbol{c}^\top x \nonumber \\
\text{s.t. } & \boldsymbol{A} x \le \boldsymbol{b}, \nonumber \\
& x_i = \hat{x}_i(\mathcal{P}), && \forall i \in \mathcal{B}_f \setminus \mathcal{U}_{f}(\mathcal{P}), \nonumber \\ 
& x_i = 0 && \forall i \in \mathcal{B}_b \setminus \mathcal{U}_{f}(\mathcal{P}): \hat{z}_i(\mathcal{P}) = 0 \nonumber \\ 
& l_i^j \le x_i \le u_i^j, && \forall i \in \mathcal{B}_b, j = \widehat{d}_i(\mathcal{P}) \nonumber \\
& x_i \in \mathbb{Z}, && \forall i \in \mathcal{I}, \nonumber \\
& x_i \in \mathbb{R}, && \forall i \notin \mathcal{I}, \nonumber
\end{align}

\noindent
Techniques for selecting \(\mathcal{U}_{f}(\mathcal{P})\) include greedy heuristics based on infeasibility or constraint violations, solver sensitivity analysis using dual or reduced-cost information, and ML models predicting which fixed variables are likely problematic.

\subsection{Bound Adjustment}\label{sec:bound_adjustment}

For wide-range backdoor variables predicted as active (non-zero), the predicted intervals may be too restrictive. The bound adjustment procedure expands the intervals for a subset $\mathcal{U}_{b}(\mathcal{P})$ of variables, while leaving the remaining variables unchanged. Let \([\tilde{l}_i^j, \tilde{u}_i^j]\) denote the adjusted bounds for \(i \in \mathcal{U}_{b}\). The bound-adjusted problem is then

\allowdisplaybreaks
\begin{align}
\phi^{ba}(\mathcal{P}) = 
\argmin_{x} \;& \boldsymbol{c}^\top x \nonumber \\
\text{s.t. } & \boldsymbol{A} x \le \boldsymbol{b}, \nonumber \\
& x_i = \hat{x}_i(\mathcal{P}), && \forall i \in \mathcal{B}_f, \nonumber \\ 
& x_i = 0 && \forall i \in \mathcal{B}_b: \hat{z}_i(\mathcal{P}) = 0 \nonumber \\ 
& l_i^j \le x_i \le u_i^j, && \forall i \in \mathcal{B}_b \setminus \mathcal{U}_{b}(\mathcal{P}), j = \widehat{d}_i(\mathcal{P}) \nonumber \\
& \tilde{l}_i^j \le x_i \le \tilde{u}_i^j, && \forall i \in \mathcal{U}_{b}(\mathcal{P}), j = \widehat{d}_i(\mathcal{P}) \nonumber \\
& x_i \in \mathbb{Z}, && \forall i \in \mathcal{I}, \nonumber \\
& x_i \in \mathbb{R}, && \forall i \notin \mathcal{I}, \nonumber
\end{align}

\noindent
Common strategies for selecting \(\mathcal{U}_{b}(\mathcal{P})\) and adjusting bounds include threshold-based relaxation using historical solution statistics, solver-guided iterative expansion until feasibility is achieved, and learning-based methods that predict effective bound expansions.

If both selective relaxation and bound adjustment are applied, the resulting reduced problem incorporates both modifications simultaneously, with variables in \(\mathcal{U}_{f}(\mathcal{P})\) relaxed and variables in \(\mathcal{U}_{b}(\mathcal{P})\) having their bounds updated, yielding a combined correction model that preserves feasibility while leveraging the ML predictions.

\section{Experimental Design}\label{sec:experiments}

This section describes the general characteristics of the test instances, the computational setting, and implementation details.

\subsection{Case Studies}

The first case study is drawn from the middle-mile consolidation network design problem (MMCNP). The MMCNP focuses on planning transportation capacity from vendors to fulfillment centers (FCs) and last-mile delivery (LMDs) centers, within required lead times. Shipments must be routed and consolidated into scheduled loads to ensure efficient delivery. The planner must decide which connections to operate, how much capacity to allocate, and how to group shipments so all demand is delivered on time. The hard setting includes 50 facilities, 2{,}000 paths, 400 arcs, and 300 commodities, and the very-hard problem setting includes 200 facilities, 24{,}000 paths, 1{,}000 arcs, and 6{,}000 commodities. In the hard setting, instances are solved on average within 18 hours. In the very-hard setting, instances do not converge and exhibit an average gap of 5.12\% after 24 hours, underscoring the difficulty of these models. Domain-specific features for the MMCNP case include: number of facilities (vendors, FCs, and LMDs), number of commodities to be transported, total demand volume aggregated over all commodities, number of direct transportation paths in the network, and the demand-to-capacity ratio defined as total demand divided by total available transportation capacity. Phase I reveals that the backdoor variables in the MMCNP case are the integer variables that capture the number of trucks flowing on each arc. These variables span a small range. The reader is referred to the original paper for a detailed description of the model and industrial context~\citep{greening2023lead,huang2024distributional}.

The second case is drawn from the strategic locomotive assignment problem (SLAP). The SLAP determines a weekly, repeatable plan for assigning locomotives to a fixed train schedule. The problem specifies the required inputs, operational constraints, and cost components governing locomotive movements and work events such as pickups and set-outs. Its main outputs are the weekly locomotive flows across the network and the event locations. These strategic decisions, including fleet sizing and event placement, provide the foundation for subsequent tactical and operational planning. The hard problem setting includes 1{,}000 nodes and 40,000 arcs, and the very-hard problem setting includes 8{,}000 nodes and 50,000 arcs. In the hard setting, instances reach optimality on average in five hours. In the very-hard setting, instances are more demanding and do not close the gap even after 24 hours, with an average gap of 0.03\%. Domain-specific features for the SLAP include: the number of locomotives, the number of arcs, the number of trains, the length of the planning horizon (number of periods), the total initial locomotive inventory, and the average number of locomotives per train. Phase I reveals that the backdoor variables in the SLAP case are the integer variables that capture the number of locomotives on each arc. These variables span a small range. The reader is referred to the original paper for a detailed description of the industrial context~\citep{kim2025practice}.

The third case is drawn from a course scheduling problem (COURSE) at Cornell University. The COURSE determines an optimal assignment of university courses to discrete time blocks. The problem specifies the required inputs and operational constraints, such as block capacities and penalty components governing scheduling conflicts between co-enrolled courses. Its main outputs are the exact placements of courses into time slots. The hard problem setting includes 200 courses and 12 time slots, and the very-hard problem setting includes 400 courses and 14 time slots. As the number of courses increases, the density of conflict constraints grows significantly, making the resulting MIPs substantially more challenging to solve. Domain-specific features for the COURSE case include: number of courses, number of time slots, total number of conflict pairs (co-enrollments), average conflict degree per course, maximum conflict degree, total penalty weight aggregated over all conflicts, average penalty per conflict, and conflict density (ratio of realized conflicts to all possible course pairs). Phase I reveals that the backdoor variables are the binary variables indicating the assignment of each course to a specific time slot. Since each course must be placed into exactly one time slot out of all available options, all remaining assignment variables for that course are strictly zero. The reader is referred to the original paper for a detailed description of the industrial context~\citep{ye2025cornell}.

The fourth case study (OCP) is the downstream supply chain of the OCP Group at the Jorf site in Morocco, one of the world’s largest phosphate-processing and export facilities. OCP is a large-scale mixed-integer linear programming model that integrates production, storage, and distribution decisions over a monthly planning horizon with daily resolution. The model captures the physical flow of materials across a highly interconnected network of conveyors, pipelines, storage units, and quays, while accounting for a rapidly expanding product portfolio and a demand-driven order fulfillment process. Key decisions include selecting shipments and loading schedules, developing daily production plans across multiple processing units, and determining implied inventory and changeover trajectories. The \emph{hard} setting includes 61 vessels, a 32-day horizon (summer), and a demand of around 1{,}066{,}290 tonnes, and the \emph{very-hard} setting includes 58 vessels, a 30-day horizon (winter), and a demand of around 1{,}797{,}910 tonnes. In the hard setting, instances are solved on average within 1000 seconds. In the very-hard setting, instances do not converge and exhibit an average gap of 2\% after 3 hours, underscoring the difficulty of these models. Domain-specific features for the OCP case include: number of vessels in the planning horizon, total demand volume (in tonnes) aggregated over all products and vessels, number of products considered in the instance, total available storage capacity across all storage units, average production capacity utilization (ratio of demand to nominal production capacity), and seasonal indicator (e.g., month or quarter of the year). Phase I reveals that the backdoor variables for the OCP case are the binary variables assigning the vessels to quays for loading over a specific period. A detailed description of the model and industrial context is available in \citep{erraqabi5}.

The fifth case study considers a multi-period, multi-product order fulfillment problem (OFP) arising in distribution and warehousing operations. The planner must decide which customer orders to fulfill, how much to ship from each warehouse in each period, and when to activate warehouse facilities to maximize total profit, defined as revenues minus handling, transportation, and setup costs. The model incorporates inventory balance constraints, warehouse capacity limits, and order-specific delivery time windows, resulting in a large-scale mixed-integer linear program. The hard problem setting includes 80 periods, 100 orders, 6 locations, and 6 products, and the very-hard problem setting includes 100 periods, 200 orders, 10 locations, and 10 products. In the hard setting, instances are solved on average within 60 minutes. In the very-hard setting, instances are solved on average within 3 hours. Domain-specific features for the OFP include: the number of customer orders, the number of warehouse locations, the number of products, the length of the planning horizon (number of periods), the total initial inventory aggregated across all products and warehouses, and the average order time-window width measuring delivery flexibility. Phase I reveals that the backdoors for the OFP case are the integer variables that capture the quantities to produce for each product in each period and each facility. These variables span a large range. The reader is referred to the original paper for a detailed description of the model and industrial context~\citep{ye2025deep}.

\begin{table}[t!]
\centering
\caption{Instances}
 \begin{tabular}{l|r|r|r|r|r}
\toprule
    Problem & Constraints & Continuous & Binary & Integer & Avg. Zeros \\
\midrule
MMCNP Hard      & 643     & - & 2{,}540  & 357      & 87\% \\
MMCNP Very-Hard & 4{,}361 & - & 23{,}984 & 1{,}135  & 85\% \\
\hline
SLAP Hard      & 80{,}982 & - & 8      & 77{,}946 & 98\% \\
SLAP Very-Hard & 82{,}996 & - & 20     & 77{,}032 & 95\%   \\
\hline
COURSE Hard      & 30{,}173 & - & 43{,}488 & 173 & 95\%\\
COURSE Very-Hard & 53{,}205 & - & 79{,}968 & 229 & 90\% \\
\hline
OCP Hard      & 772{,}383     & 925{,}502 & 11{,}155  & -      & 96\% \\
OCP Very-Hard & 370{,}827  & 435{,}536 & 15{,}237 & -  & 92\% \\
\hline
OFP Hard      & 143{,}892 & 2{,}880 & 580 & 288{,}580 & 90\%  \\
OFP Very-Hard & 1{,}012{,}900 & 10{,}000 & 1{,}200 & 2{,}001{,}200 & 91\% \\
\bottomrule
\end{tabular}
\label{tab:instances}
\end{table}

\begin{table}[t!]
\centering
\caption{Case Studies Classification based on the Taxonomy}
\label{tab:case studies classification}
\begin{adjustbox}{width=\textwidth}
\begin{tabular}{l|cc|cc|cc|cc|cc}
\toprule
\multirow{2}{*}{Taxonomy} & \multicolumn{2}{c|}{MMCNP} & \multicolumn{2}{c|}{SLAP} & \multicolumn{2}{c|}{COURSE} & \multicolumn{2}{c|}{OCP} & \multicolumn{2}{c}{OFP}\\
\cline{2-11}
                      & Hard      & Very-Hard   & Hard       & Very-Hard    & Hard      & Very-Hard & Hard      & Very-Hard & Hard      & Very-Hard   \\
\midrule
Problem Size          & $P_1$        & $P_1$          &   $P_1$         & $P_1$           & $P_1$        & $P_1$     &   $P_2$         & $P_2$  & $P_2$         & $P_2$  \\
Variable Type         & $V_1$        & $V_1$          &  $V_1$         & $V_1$           & $V_1$        &  $V_1$  & $V_1$          &  $V_1$      & $V_2$          &  $V_2$    \\
Feature Type          & $F_1$ / $F_2$   & $F_1$ / $F_2$     & $F_1$ / $F_2$    & $F_1$ / $F_2$      & $F_1$ / $F_2$   & $F_1$ / $F_2$ & $F_1$ / $F_2$    & $F_1$ / $F_2$ & $F_1$ / $F_2$    & $F_1$ / $F_2$ \\
\bottomrule
\end{tabular}
\end{adjustbox}
\end{table}
Table~\ref{tab:instances} overviews instances in each setting for each case study and reports the number of constraints, the number of continuous variables, the number of binary variables, the number of integer variables, and the average percentage of zero-valued backdoor variables in the optimal solution. These case studies are well-suited to evaluating the proposed \pop{} framework, as they represent various cases discussed in the taxonomy as summarized in Table~\ref{tab:case studies classification}. Throughout this paper, problems with fewer than 100,000 backdoor variables are classified as medium-scale ($P_1$), whereas problems with at least 100,000 backdoor variables are classified as large-scale ($P_2$). This threshold was selected empirically based on the scalability of the learning models and serves as a guideline for choosing between global and variable-wise prediction architectures.

\subsection{Experimental Setting}

\textbf{Data Generation.} For each case study, training and test instances are constructed from two reference problem instances that represent different levels of difficulty, referred to as \emph{hard} and \emph{very-hard}. These reference instances are obtained from realistic operational data settings and serve as representative baselines for the underlying optimization problems.

Throughout the experiments, all generated instances preserve the structural formulation of the reference problem, while demand parameters are perturbed to create diverse operational scenarios following a procedure inspired by \cite{akhlaghi2025propel}. Demand uncertainty in real-world optimization problems typically exhibits both positive correlations, driven by factors such as seasonality and market trends, and negative correlations arising from substitution effects or resource competition. The data generation process aims at capturing these characteristics while maintaining consistency with observed operational patterns. The perturbed datasets reflect realistic operational variability and produce integer variables with diverse activation patterns and wide value ranges, making them well-suited for evaluating the learning and correction components of the \pop{} framework.

\noindent \textbf{Computational Setting.} For each case study, a total of 600 MIP instances were generated following the procedure described above. These instances were split into 400 training instances, 100 validation instances, and 100 testing instances. The training set was used to fit the supervised learning models during Phase II, the validation set was used for hyperparameter tuning and model selection, and the testing set was reserved exclusively for performance evaluation.

The DNNs used for both classification and interval prediction in Phase II are fully connected multilayer perceptrons with ReLU activation functions. Hyperparameters were selected via grid search over learning rates $\{0.001, 0.005\}$, number of hidden layers $\{3,4\}$, and hidden layer dimensions $\{32,64,128\}$. The number of instance-level and variable-level features used determines the input dimension of each network. All models were trained for 100 epochs with a batch size of 32 using the Adam optimizer. Model selection was performed based on validation performance, and final results are reported on held-out test sets.

Classification models were evaluated using standard metrics such as F1 score and confusion matrices, while interval prediction models were assessed using classification accuracy over the predicted buckets. For medium-scale problems ($P_1$), a single global model was trained for both classification and interval prediction tasks. In contrast, for large-scale problems ($P_2$), separate variable-wise models were trained for each backdoor variable.

All learning models were implemented in Python using PyTorch. Experiments were conducted using Gurobi Optimizer version~13.0.0. All optimization runs were executed on a machine equipped with an Intel\textsuperscript{\textregistered} Core\textsuperscript{TM} i7--8700 processor operating at 3.20\,GHz, with 64\,GiB of system memory, running Oracle Linux Server release~7.7. 

The reduced and corrected MIP models were solved using 32 CPU threads. The termination criterion was defined by setting the Gurobi parameter \texttt{MIPGap} to 0.5\%, and \texttt{MIPFocus} was set to 1 to encourage rapid identification of high-quality feasible solutions. All reported runtimes correspond to real (wall-clock) time. Unless otherwise stated, all reported objective values correspond to the best feasible solution returned by the solver within the allotted time limit.

\noindent \textbf{Metrics.} It is worth highlighting that strong predictive accuracy does not necessarily translate into high-quality optimization decisions \citep{elmachtoub2020decision}. Accordingly, the evaluation focuses exclusively on the quality of solutions obtained by solving the reduced and corrected MIP models produced by the \pop{} framework, rather than on standalone ML performance metrics.

Solution quality is measured using two standard optimization metrics. The \emph{primal gap} (PG) is defined as the relative difference between the objective value of the best feasible solution found by a method and the best known bound, evaluated at termination. This metric captures the final solution quality achieved within the computational budget. Formally, let $z^\text{primal}(t)$ denote the objective value of the best feasible solution found up to time $t$, and let $z^\text{bound}(t)$ denote the best available dual bound at time $t$. At termination time $T$, the \emph{primal gap} is defined as
\[
\text{PrimalGap}(T) 
:= 
\frac{\big| z^\text{primal}(T) - z^\text{bound}(T) \big|}
{\max\{ |z^\text{primal}(T)|, \epsilon \}},
\]
where $\epsilon > 0$ is a small constant introduced to avoid division by zero.

In addition, the results report the \emph{primal integral} (PI), which measures the time-integrated primal gap over the entire solve horizon, thereby accounting for both solution quality and the speed at which high-quality feasible solutions are identified. Together, these metrics provide a comprehensive assessment of the decision-making effectiveness of the \pop{} framework, capturing not only the quality of final solutions but also the temporal behavior of the optimization process relative to the baseline. Formally:
\[
\text{PrimalIntegral}(T)
:=
\int_{0}^{T} 
\frac{\big| z^\text{primal}(t) - z^\text{bound}(t) \big|}
{\max\{ |z^\text{primal}(t)|, \epsilon \}}
\, dt.
\]

\noindent \textbf{Comparison.} The \pop{} framework is benchmarked against state-of-the-art solution approaches to assess its ability to deliver high-quality solutions under limited computational budgets. This comparison highlights the trade-off between solution quality and runtime and illustrates how learning-augmented optimization can outperform conventional solvers on large-scale and computationally challenging instances. The considered methods are:
\begin{itemize}
    \item \textbf{MIP:} Directly solving the full model using a state-of-the-art MIP solver, i.e., Gurobi with default parameter settings.
    \item \textbf{\textsc{PROPEL}:} The learning-augmented optimization framework proposed in \citet{akhlaghi2025propel}.
    \item \textbf{\textsc{ID-PaS+}:} The identity aware predict-and-search introduced in \citet{cai2025id}.
\end{itemize}

All methods are evaluated under a runtime limit of 600 seconds over the five cases: OCP, MMCNP, OFP, SLAP, and COURSE. For each time budget, solution quality is assessed using the PG at termination and the PI over the entire solve horizon, thereby jointly evaluating final solution quality and the speed at which high-quality feasible solutions are obtained. ID-PaS+ does not apply to OFP because its methodology is restricted to bounded-domain variables and does not support the wide-range integer variables present in OFP.

\section{Computational Results}\label{sec:results}

This section evaluates the computational effectiveness of the proposed \pop{} framework through an extensive experimental study. First, the \pop{} performance is assessed relative to state-of-the-art approaches. Then, the computational performance of each \pop{} phase is quantified. Last, several managerial insights relevant to large-scale decision-making contexts are highlighted. The detailed backdoor identification (Phase 1) is provided in Appendix \ref{A}.

\subsection{Comparison with State-of-Art}\label{sec:overall_performance}

Table \ref{table:performance_comparison} shows the PG and PI averaged over 100 test instances for each benchmark, and reports the mean, standard deviation, and the number of instances each method wins. The best-performing entries are highlighted in bold for clarity.

\begin{table}[t!]
\centering
\caption{Performance comparison of baseline and proposed methods across benchmark instances in terms of PG and PI. Results are reported as mean, standard deviation, and number of best-performing instances (Wins) over 100 instances per benchmark. Lower values indicate better performance.}
\begin{adjustbox}{width=0.8\textwidth}
\begin{tabular}{l|l|ccc|ccc}
\toprule
        \multirow{2}{*}{Problem}                &     \multirow{2}{*}{Method}              & \multicolumn{3}{c|}{PG (\%) $\downarrow$}                   & \multicolumn{3}{c}{PI  $\downarrow$}                      \\ \cline{3-8}
            &     & Mean            & Std Dev         & Wins        & Mean          & Std Dev       & Wins         \\ \midrule
\multirow{3}{*}{MMCNP Hard}  & MIP           & 0.16          & 0.25          & 10           & 3.30         & 3.28          & 0            \\
                       & \textsc{PROPEL}      & 0.09         & 0.15 & 12         & 2.12          & 2.22         & 6           \\
                       & \textsc{ID-PaS+} & 0.07 & 0.17          & 18 & 1.94 & 1.99 & 10  \\
                       &  \pop{} & \textbf{0.05}         & \textbf{0.10} & \textbf{60}         & \textbf{0.51}          & \textbf{1.45}         & \textbf{84}           \\ \midrule
\multirow{3}{*}{MMCNP Very-Hard} & MIP            & 1.05          & 0.49          & 2          & 35.11         & 5.91          & 0            \\
                       & \textsc{PROPEL}      & 0.15        & 0.27 & 14         & 12.33          & 2.45         & 9           \\
                       & \textsc{ID-PaS+} & 0.12 & 0.23 & 14          & 11.27          & 2.67 & 9          \\
                       &  \pop{} & \textbf{0.04}         & \textbf{0.08} & \textbf{70}         & \textbf{2.70}          & \textbf{0.67}         & \textbf{82}           \\ \midrule
\multirow{3}{*}{SLAP Hard} & MIP            & 0.049           & 0.054 & 9          & 1.36          & 0.59         & 0           \\
                       & \textsc{PROPEL}      & 0.028  & 0.041          & 10         & 1.07 & 0.25         & 8  \\ 
                       & \textsc{ID-PaS+} & 0.024  & 0.032          & 13         &1.16 & 0.43         & 7  \\ 
                       &  \pop{} & \textbf{0.011}  &  \textbf{0.014}          &  \textbf{68}         & \textbf{0.42} &  \textbf{0.16}         & \textbf{85}           \\ \midrule               
\multirow{3}{*}{SLAP Very-Hard} & MIP            & 0.18          & 0.21         & 7          & 7.85         & 2.12          & 0            \\
                       & \textsc{PROPEL}      & 0.15  & 0.33          & 6         & 6.12 & 1.18         & 10  \\ 
                       & \textsc{ID-PaS+} & 0.13 & 0.15 & 11 & 5.36 & 1.55 & 10  \\
                       &  \pop{} & \textbf{0.08}  &  \textbf{0.07}          &  \textbf{76}         & \textbf{1.24} &  \textbf{0.44}         & \textbf{80}           \\ \midrule
\multirow{4}{*}{COURSE Hard} & MIP            & 0.023           & 0.010 & 2          & 1.97          & 1.78        & 1          \\
                        & \textsc{PROPEL} &  0.013 & 0.008 & 18 & 1.90 & 1.59 & 3\\
                       & \textsc{ID-PaS+} & 0.004  & 0.003          & 25          & 1.75 & 1.71         & 5  \\ 
                       &  \pop{}      & \textbf{0.004}           & \textbf{0.002}         & \textbf{55} & \textbf{1.43} & \textbf{1.55} & \textbf{91}         \\                       \midrule 
\multirow{4}{*}{COURSE Very-Hard} & MIP           & 0.26         & 0.44         & 3          & 9.87         & 6.35          & 0            \\
                    & \textsc{PROPEL} & 0.13 & 0.31 & 6 & 9.14 & 5.03 & 3\\
                    & \textsc{ID-PaS+} & 0.07 & 0.23 & 31 & 7.61 & 4.56 & 10  \\
                    &  \pop{}      & \textbf{0.05}          & \textbf{0.11}          & \textbf{60}          & \textbf{6.76}          & \textbf{4.22}         & \textbf{87}           \\ \midrule
\multirow{3}{*}{OCP Hard}  & MIP           & 0.09          &  0.42          & 0           & 44.53         & 8.33          &  0           \\
                       & \textsc{PROPEL}      & 0.08          & 0.13          & 15           & 11.72         & 2.97          & 7            \\
                       & \textsc{ID-PaS+} & 0.08          & 0.08          & 20           & 11.13         & 3.11          & 8            \\ 
                       &  \pop{} & \textbf{0.07}         & \textbf{0.05}          & \textbf{65}           & \textbf{3.21}         &  \textbf{1.56}          & \textbf{85}            \\                   \midrule
\multirow{3}{*}{OCP Very-Hard} & MIP           & 2.61          & 0.78          & 3           & 41.78         &  6.41          &   2          \\
                       & \textsc{PROPEL}      & 1.72          & 0.23          & 16           & 9.94         & 3.42          & 10            \\
                       & \textsc{ID-PaS+} & 1.98          & 0.15          & 10           & 9.28         & 3.55          & 8            \\ 
                       &  \pop{} & \textbf{0.80}          &   \textbf{0.03}          & \textbf{71}           & \textbf{3.79}         & \textbf{1.22}          & \textbf{80}            \\   \midrule
\multirow{3}{*}{OFP Hard} & MIP           & 0.30          & 0.41          & 4           & 4.71         & 1.32          & 0            \\
                       & \textsc{PROPEL}      & 0.20          & 0.31          & 15           & 3.12         & 0.75          & 5            \\
                       & \textsc{ID-PaS+} & -          & -          & -           & -         & -          & -            \\ 
                       &  \pop{} & \textbf{0.09}  &  \textbf{0.05}          &  \textbf{81}         & \textbf{0.79} &  \textbf{0.27}         & \textbf{95}           \\ \midrule             
\multirow{3}{*}{OFP Very-Hard} & MIP           & 1.14          & 0.25          & 3           & 8.11         &  1.12         & 0            \\
                       & \textsc{PROPEL}      & 0.78          & 0.45          & 18           & 4.44         & 1.02          & 12                 \\
                       & \textsc{ID-PaS+} & -          & -          & -           & -         & -          & -            \\ 
                       &  \pop{} & \textbf{0.13}  &  \textbf{0.08}          &  \textbf{79}         & \textbf{1.16} &  \textbf{0.25}         & \textbf{88}           \\\bottomrule
\end{tabular}
\end{adjustbox}
\label{table:performance_comparison}
\end{table}

Across all problem families and difficulty levels,  \pop{} consistently achieves the lowest mean PG and PI, often by a substantial margin. In the very-hard settings, the improvement is particularly pronounced. For example, in OCP Very-Hard,  \pop{} reduces the mean PG from 1.72 and 1.98 for \textsc{PROPEL} and \textsc{ID-PaS+}, and 2.61 for MIP, to 0.80, while simultaneously decreasing the PI to 3.79 compared to values between 9 and 42 for competing methods. This demonstrates that  \pop{} improves both final solution quality and the speed at which high-quality incumbents are identified. In hard settings, all learning-based methods remain competitive, yet  \pop{} still achieves the smallest PG and PI, and exhibits lower variability. This indicates not only improved solution quality but also greater stability across instances. The reduction in PI is especially notable, showing that  \pop{} consistently identifies strong feasible solutions early in the search process. A Wilcoxon signed-rank test ($p < 0.01$)~\citep{woolson2007wilcoxon} confirms that all reported performance differences between  \pop{} and the baselines are statistically significant, validating the robustness of the observed improvements across the various cases.

From an algorithmic perspective, the distinction between predicting zeros among all integer variables and predicting both zeros and non-zeros on a selected subset of backdoor variables reflects two fundamentally different mechanisms for reducing combinatorial complexity. Predicting zeros across the full set of integer variables, as done in \textsc{PROPEL} and \textsc{ID-PaS+}, follows a conservative pruning strategy. By identifying likely inactive variables, this strategy reduces the search space while preserving flexibility. The solver determines the remaining assignments. This strategy exploits sparsity, which is common in many integer programs, and can substantially reduce the feasible region when zero predictions are accurate. However, it provides limited structural guidance, since the solver must still determine which of the remaining variables should take positive values.

In contrast, predicting both zeros and non-zeros on a targeted subset of backdoor variables, as in  \pop{}, directly addresses the structural drivers of difficulty. Rather than distributing predictions across the entire integer space, this strategy concentrates on variables that encode the core combinatorial decisions. By fixing both inactive and active assignments within this subset, the method not only prunes the search space but also shapes it, thereby reducing the effective combinatorial burden faced by the solver.

The performance trends in Table~\ref{table:performance_comparison} align with this intuition. As instance difficulty increases, the advantage of  \pop{} becomes more pronounced in both PG and PI. This pattern suggests that when complexity is driven by a relatively small number of structural decisions, accurately predicting their values yields stronger simplification than broadly predicting zeros across all integer variables. In easier instances, where sparsity alone may already provide sufficient reduction, the PG between strategies is smaller, consistent with the more moderate differences observed in the table. Overall, the relative gains of  \pop{} increase with problem hardness, indicating that ML and structured parameter selection provide increasingly effective guidance in large and complex search spaces.

\subsection{Computational Performance of \pop{}}\label{sec:benefits_phases}

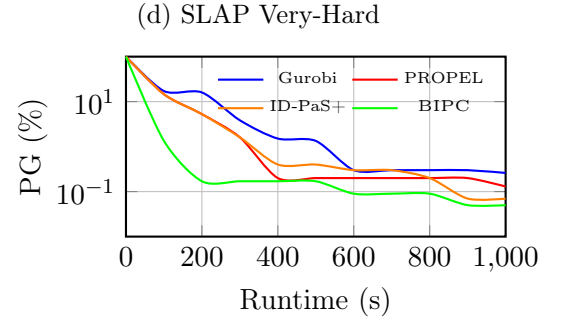
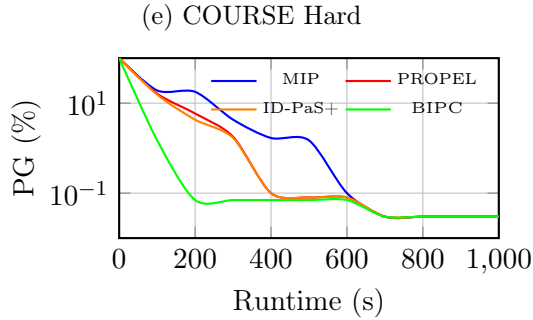
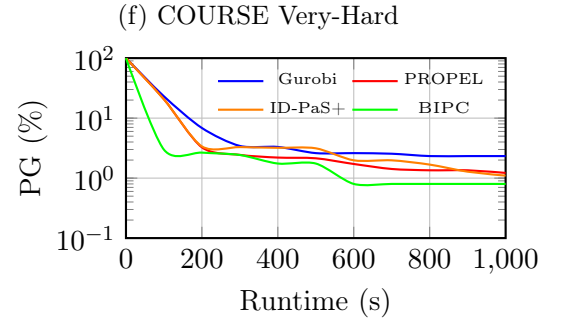
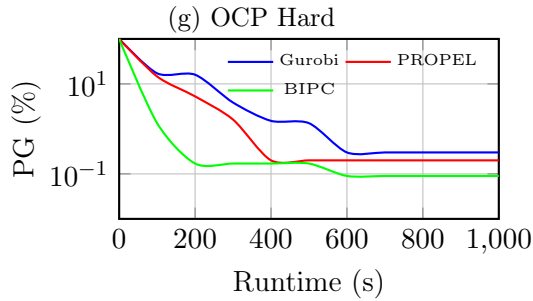
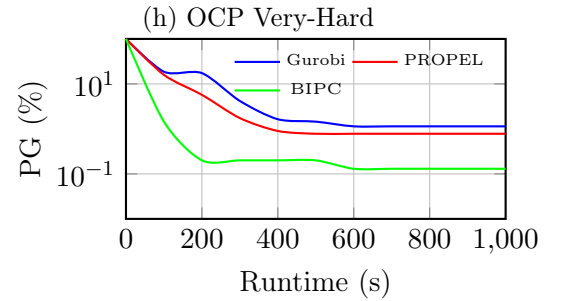
\begin{figure}[t!]
    \centering

\begin{subfigure}{0.4\textwidth}
    \centering
    \begin{tikzpicture}
        \begin{semilogyaxis}[
            width=\textwidth,
            height=0.6\textwidth,
            xlabel={Runtime (s)},
            ylabel={PG (\%)},
            xmin=0, xmax=1000,
            ymin=0.01, ymax=100,
            grid=major,
            thick,
            smooth,
            legend style={
                font=\tiny,
                at={(0.98,0.98)},
                anchor=north east,
                draw=none,
                fill=none,
                row sep=1pt
            },
            legend columns=2
        ]

        \addplot [color=blue] coordinates {
            (0,100.00) (100,15.31) (200,14.26) (300,3.43)
            (400,1.34) (500,1.19) (600,0.16) (700,0.16)
            (800,0.16) (900,0.16) (1000,0.16)
        };
        \addlegendentry{Gurobi}

        \addplot [color=red] coordinates {
            (0,100.00) (100,13.04) (200,4.70) (300,1.43)
            (400,0.18) (500,0.16) (600,0.09) (700,0.09)
            (800,0.09) (900,0.09) (1000,0.09)
        };
        \addlegendentry{PROPEL}

        \addplot [color=orange] coordinates {
            (0,100.00) (100,12.35) (200,3.43) (300,1.34)
            (400,0.09) (500,0.08) (600,0.07) (700,0.07)
            (800,0.07) (900,0.07) (1000,0.07)
        };
        \addlegendentry{ID-PaS+}

        \addplot [color=green] coordinates {
            (0,100.00) (100,1.19) (200,0.06) (300,0.06)
            (400,0.06) (500,0.06) (600,0.05) (700,0.05)
            (800,0.05) (900,0.05) (1000,0.05)
        };
        \addlegendentry{ \pop{}}

        \end{semilogyaxis}
    \end{tikzpicture}
    \caption{MMCNP Hard}
    \label{fig:pg_runtime_all_methods_MMCNP_hard}
\end{subfigure}
    \hfill
\begin{subfigure}{0.4\textwidth}
    \centering
    \begin{tikzpicture}
        \begin{semilogyaxis}[
            width=\textwidth,
            height=0.6\textwidth,
            xlabel={Runtime (s)},
            ylabel={PG (\%)},
            xmin=0, xmax=1000,
            ymin=0.01, ymax=100,
            grid=major,
            thick,
            smooth,
            legend style={
                font=\tiny,
                at={(0.98,0.98)},
                anchor=north east,
                draw=none,
                fill=none,
                row sep=1pt
            },
            legend columns=2
        ]

        \addplot [color=blue] coordinates {
            (0,100.00) (100,18.3744) (200,17.1168) (300,4.1184)
            (400,1.6128) (500,1.4304) (600,1.05) (700,1.05)
            (800,1.05) (900,1.05) (1000,1.05)
        };
        \addlegendentry{Gurobi}

        \addplot [color=red] coordinates {
            (0,100.00) (100,15.648) (200,5.6448) (300,1.7184)
            (400,0.196) (500,0.1768) (600,0.15) (700,0.15)
            (800,0.15) (900,0.15) (1000,0.15)
        };
        \addlegendentry{PROPEL}

        \addplot [color=orange] coordinates {
            (0,100.00) (100,14.8224) (200,4.1184) (300,1.6128)
            (400,0.196) (500,0.1768) (600,0.12) (700,0.12)
            (800,0.12) (900,0.12) (1000,0.12)
        };
        \addlegendentry{ID-PaS+}

        \addplot [color=green] coordinates {
            (0,100.00) (100,1.4304) (200,0.0672) (300,0.0672)
            (400,0.0672) (500,0.0672) (600,0.04) (700,0.04)
            (800,0.04) (900,0.04) (1000,0.04)
        };
        \addlegendentry{ \pop{}}

        \end{semilogyaxis}
    \end{tikzpicture}
    \caption{MMCNP Very-Hard}
    \label{fig:pg_runtime_all_methods_MMCNP_very_hard}
\end{subfigure}

\begin{subfigure}{0.4\textwidth}
    \centering
    \begin{tikzpicture}
        \begin{semilogyaxis}[
            width=\textwidth,
            height=0.6\textwidth,
            xlabel={Runtime (s)},
            ylabel={PG (\%)},
            xmin=0, xmax=1000,
            ymin=0.01, ymax=100,
            grid=major,
            thick,
            smooth,
            no markers,
            legend style={
                font=\tiny,
                at={(0.98,0.98)},
                anchor=north east,
                draw=none,
                fill=none,
                row sep=1pt
            },
            legend columns=2
        ]

        \addplot [color=blue] coordinates {
            (0,100.00) (100,18.37) (200,17.12) (300,4.12)
            (400,1.61) (500,1.43) (600,0.05) (700,0.05)
            (800,0.05) (900,0.05) (1000,0.05)
        };
        \addlegendentry{Gurobi}

        \addplot [color=red] coordinates {
            (0,100.00) (100,15.65) (200,5.64) (300,1.72)
            (400,0.10) (500,0.08) (600,0.03) (700,0.03)
            (800,0.03) (900,0.03) (1000,0.03)
        };
        \addlegendentry{PROPEL}

        \addplot [color=orange] coordinates {
            (0,100.00) (100,14.82) (200,4.12) (300,1.61)
            (400,0.10) (500,0.08) (600,0.02) (700,0.02)
            (800,0.02) (900,0.02) (1000,0.02)
        };
        \addlegendentry{ID-PaS+}

        \addplot [color=green] coordinates {
            (0,100.00) (100,1.43) (200,0.07) (300,0.07)
            (400,0.07) (500,0.07) (600,0.01) (700,0.01)
            (800,0.01) (900,0.01) (1000,0.01)
        };
        \addlegendentry{ \pop{}}

        \end{semilogyaxis}
    \end{tikzpicture}
    \caption{SLAP Hard}
    \label{fig:pg_runtime_all_methods_SLAP_hard}
\end{subfigure}
    \hfill
\begin{subfigure}{0.4\textwidth}
    \centering
    \begin{tikzpicture}
        \begin{semilogyaxis}[
            width=\textwidth,
            height=0.6\textwidth,
            xlabel={Runtime (s)},
            ylabel={PG (\%)},
            xmin=0, xmax=1000,
            ymin=0.01, ymax=100,
            grid=major,
            thick,
            smooth,
            no markers,
            legend style={
                font=\tiny,
                at={(0.98,0.98)},
                anchor=north east,
                draw=none,
                fill=none,
                row sep=1pt
            },
            legend columns=2
        ]

        \addplot [color=blue] coordinates {
            (0,100.00) (100,15.62) (200,14.55) (300,3.50)
            (400,1.37) (500,1.22) (600,0.18) (700,0.18)
            (800,0.18) (900,0.18) (1000,0.18)
        };
        \addlegendentry{Gurobi}

        \addplot [color=red] coordinates {
            (0,100.00) (100,13.30) (200,4.80) (300,1.46)
            (400,0.18) (500,0.17) (600,0.15) (700,0.15)
            (800,0.15) (900,0.15) (1000,0.15)
        };
        \addlegendentry{PROPEL}

        \addplot [color=orange] coordinates {
            (0,100.00) (100,12.60) (200,3.50) (300,1.37)
            (400,0.18) (500,0.17) (600,0.13) (700,0.13)
            (800,0.13) (900,0.13) (1000,0.13)
        };
        \addlegendentry{ID-PaS+}

        \addplot [color=green] coordinates {
            (0,100.00) (100,1.22) (200,0.16) (300,0.16)
            (400,0.16) (500,0.16) (600,0.08) (700,0.08)
            (800,0.08) (900,0.08) (1000,0.08)
        };
        \addlegendentry{ \pop{}}

        \end{semilogyaxis}
    \end{tikzpicture}
    \caption{SLAP Very-Hard}
    \label{fig:pg_runtime_all_methods_SLAP_very_hard}
\end{subfigure}

\begin{subfigure}{0.4\textwidth}
    \centering
    \begin{tikzpicture}
        \begin{semilogyaxis}[
            width=\textwidth,
            height=0.6\textwidth,
            xlabel={Runtime (s)},
            ylabel={PG (\%)},
            xmin=0, xmax=1000,
            ymin=0.001, ymax=100,
            grid=major,
            thick,
            smooth,
            no markers,
            legend style={
                font=\tiny,
                at={(0.98,0.98)},
                anchor=north east,
                draw=none,
                fill=none,
                row sep=1pt
            },
            legend columns=2
        ]

        \addplot [color=blue] coordinates {
            (0,100.00) (100,18.37) (200,17.12) (300,4.12)
            (400,1.61) (500,1.43) (600,0.05) (700,0.05)
            (800,0.023) (900,0.023) (1000,0.023)
        };
        \addlegendentry{Gurobi}

        \addplot [color=red] coordinates {
            (0,100.00) (100,15.65) (200,5.64) (300,1.72)
            (400,0.10) (500,0.08) (600,0.03) (700,0.03)
            (800,0.013) (900,0.013) (1000,0.013)
        };
        \addlegendentry{PROPEL}

        \addplot [color=orange] coordinates {
            (0,100.00) (100,14.82) (200,4.12) (300,1.61)
            (400,0.10) (500,0.08) (600,0.01) (700,0.01)
            (800,0.01) (900,0.004) (1000,0.004)
        };
        \addlegendentry{ID-PaS+}

        \addplot [color=green] coordinates {
            (0,100.00) (100,1.43) (200,0.07) (300,0.07)
            (400,0.07) (500,0.07) (600,0.01) (700,0.01)
            (800,0.004) (900,0.004) (1000,0.004)
        };
        \addlegendentry{ \pop{}}

        \end{semilogyaxis}
    \end{tikzpicture}
    \caption{COURSE Hard}
    \label{fig:pg_runtime_all_methods_COURSE_hard}
\end{subfigure}
    \hfill
\begin{subfigure}{0.4\textwidth}
    \centering
    \begin{tikzpicture}
        \begin{semilogyaxis}[
            width=\textwidth,
            height=0.6\textwidth,
            xlabel={Runtime (s)},
            ylabel={PG (\%)},
            xmin=0, xmax=1000,
            ymin=0.01, ymax=100,
            grid=major,
            thick,
            smooth,
            no markers,
            legend style={
                font=\tiny,
                at={(0.98,0.98)},
                anchor=north east,
                draw=none,
                fill=none,
                row sep=1pt
            },
            legend columns=2
        ]

        \addplot [color=blue] coordinates {
            (0,100.00) (100,17.18) (200,16.00) (300,3.85)
            (400,1.51) (500,1.34) (600,0.30) (700,0.30)
            (800,0.30) (900,0.30) (1000,0.26)
        };
        \addlegendentry{Gurobi}

        \addplot [color=red] coordinates {
            (0,100.00) (100,14.63) (200,5.28) (300,1.61)
            (400,0.20) (500,0.20) (600,0.20) (700,0.20)
            (800,0.20) (900,0.20) (1000,0.13)
        };
        \addlegendentry{PROPEL}

        \addplot [color=orange] coordinates {
            (0,100.00) (100,14.63) (200,5.28) (300,1.61)
            (400,0.40) (500,0.40) (600,0.30) (700,0.30)
            (800,0.20) (900,0.07) (1000,0.07)
        };
        \addlegendentry{ID-PaS+}

        \addplot [color=green] coordinates {
            (0,100.00) (100,1.34) (200,0.17) (300,0.17)
            (400,0.17) (500,0.17) (600,0.09) (700,0.09)
            (800,0.09) (900,0.05) (1000,0.05)
        };
        \addlegendentry{ \pop{}}

        \end{semilogyaxis}
    \end{tikzpicture}
    \caption{COURSE Very-Hard}
    \label{fig:pg_runtime_all_methods_COURSE_very_hard}
\end{subfigure}

\begin{subfigure}{0.4\textwidth}
    \centering
    \begin{tikzpicture}
        \begin{semilogyaxis}[
            width=\textwidth,
            height=0.6\textwidth,
            xlabel={Runtime (s)},
            ylabel={PG (\%)},
            xmin=0, xmax=1000,
            ymin=0.01, ymax=100,
            grid=major,
            thick,
            smooth,
            no markers,
            legend style={
                font=\tiny,
                at={(0.98,0.98)},
                anchor=north east,
                draw=none,
                fill=none,
                row sep=1pt
            },
            legend columns=2
        ]
        \addplot coordinates {
            (0,100.00) (100,19.14) (200,17.83) (300,4.29)
            (400,1.68) (500,1.49) (600,0.10) (700,0.03)
            (800,0.03) (900,0.03) (1000,0.03)
        };
        \addlegendentry{MIP}

        \addplot coordinates {
            (0,100.00) (100,16.30) (200,5.88) (300,1.79)
            (400,0.10) (500,0.08) (600,0.08) (700,0.03)
            (800,0.03) (900,0.03) (1000,0.03)
        };
        \addlegendentry{PROPEL}

        \addplot [color=orange] coordinates {
            (0,100.00) (100,15.44) (200,4.29) (300,1.68)
            (400,0.10) (500,0.08) (600,0.08) (700,0.03)
            (800,0.03) (900,0.03) (1000,0.03)
        };
        \addlegendentry{ID-PaS+}

        \addplot [color=green] coordinates {
            (0,100.00) (100,1.49) (200,0.07) (300,0.07)
            (400,0.07) (500,0.07) (600,0.07) (700,0.03)
            (800,0.03) (900,0.03) (1000,0.03)
        };
        \addlegendentry{ \pop{}}

        \end{semilogyaxis}
    \end{tikzpicture}
    \caption{OCP Hard}
    \label{fig:pg_runtime_all_methods_OCP_hard}
\end{subfigure}
  \hfill 
\begin{subfigure}{0.4\textwidth}
    \centering
    \begin{tikzpicture}
        \begin{semilogyaxis}[
            width=\textwidth,
            height=0.6\textwidth,
            xlabel={Runtime (s)},
            ylabel={PG (\%)},
            xmin=0, xmax=1000,
            ymin=0.1, ymax=100,
            grid=major,
            thick,
            smooth,
            legend style={
                font=\tiny,
                at={(0.98,0.98)},
                anchor=north east,
                draw=none,
                fill=none,
                row sep=1pt
            },
            legend columns=2
        ]

        \addplot [color=blue] coordinates {
            (0,100.00) (100,23.07) (200,6.85) (300,3.44)
            (400,3.31) (500,2.61) (600,2.61) (700,2.55)
            (800,2.33) (900,2.33) (1000,2.33)
        };
        \addlegendentry{Gurobi}

        \addplot [color=red] coordinates {
            (0,100.00) (100,20.65) (200,3.26) (300,2.44)
            (400,2.20) (500,2.14) (600,1.72) (700,1.42)
            (800,1.35) (900,1.35) (1000,1.22)
        };
        \addlegendentry{PROPEL}

        \addplot [color=orange] coordinates {
            (0,100.00) (100,20.56) (200,3.38) (300,3.29)
            (400,3.20) (500,3.14) (600,1.98) (700,1.98)
            (800,1.68) (900,1.28) (1000,1.10)
        };
        \addlegendentry{ID-PaS+}

        \addplot [color=green] coordinates {
            (0,100.00) (100,2.98) (200,2.66) (300,2.43)
            (400,1.75) (500,1.75) (600,0.80) (700,0.80)
            (800,0.80) (900,0.80) (1000,0.80)
        };
        \addlegendentry{ \pop{}}

        \end{semilogyaxis}
    \end{tikzpicture}
    \caption{OCP Very-Hard}
    \label{fig:pg_runtime_all_methods_OCP_very_hard}
\end{subfigure}

\begin{subfigure}{0.4\textwidth}
    \centering
    \begin{tikzpicture}
        \begin{semilogyaxis}[
            width=\textwidth,
            height=0.6\textwidth,
            xlabel={Runtime (s)},
            ylabel={PG (\%)},
            xmin=0, xmax=1000,
            ymin=0.01, ymax=100,
            grid=major,
            thick,
            smooth,
            no markers,
            legend style={
                font=\tiny,
                at={(0.98,0.98)},
                anchor=north east,
                draw=none,
                fill=none,
                row sep=1pt
            },
            legend columns=2
        ]

        \addplot [color=blue] coordinates {
            (0,100.00) (100,17.18) (200,16.00) (300,3.85)
            (400,1.51) (500,1.34) (600,0.30) (700,0.30)
            (800,0.30) (900,0.30) (1000,0.30)
        };
        \addlegendentry{Gurobi}

        \addplot [color=red] coordinates {
            (0,100.00) (100,14.63) (200,5.28) (300,1.61)
            (400,0.20) (500,0.20) (600,0.20) (700,0.20)
            (800,0.20) (900,0.20) (1000,0.20)
        };
        \addlegendentry{PROPEL}

        \addplot [color=green] coordinates {
            (0,100.00) (100,1.34) (200,0.17) (300,0.17)
            (400,0.17) (500,0.17) (600,0.09) (700,0.09)
            (800,0.09) (900,0.09) (1000,0.09)
        };
        \addlegendentry{ \pop{}}

        \end{semilogyaxis}
    \end{tikzpicture}
    \caption{OFP Hard}
    \label{fig:pg_runtime_all_methods_OFP_hard}
\end{subfigure}
    \hfill
\begin{subfigure}{0.4\textwidth}
    \centering
    \begin{tikzpicture}
        \begin{semilogyaxis}[
            width=\textwidth,
            height=0.6\textwidth,
            xlabel={Runtime (s)},
            ylabel={PG (\%)},
            xmin=0, xmax=1000,
            ymin=0.01, ymax=100,
            grid=major,
            thick,
            smooth,
            no markers,
            legend style={
                font=\tiny,
                at={(0.98,0.98)},
                anchor=north east,
                draw=none,
                fill=none,
                row sep=1pt
            },
            legend columns=2
        ]

        \addplot [color=blue] coordinates {
            (0,100.00) (100,18.65) (200,17.37) (300,4.18)
            (400,1.64) (500,1.45) (600,1.14) (700,1.14)
            (800,1.14) (900,1.14) (1000,1.14)
        };
        \addlegendentry{Gurobi}

        \addplot [color=red] coordinates {
            (0,100.00) (100,15.88) (200,5.73) (300,1.74)
            (400,0.90) (500,0.78) (600,0.78) (700,0.78)
            (800,0.78) (900,0.78) (1000,0.78)
        };
        \addlegendentry{PROPEL}

        \addplot [color=green] coordinates {
            (0,100.00) (100,1.45) (200,0.20) (300,0.20)
            (400,0.20) (500,0.20) (600,0.13) (700,0.13)
            (800,0.13) (900,0.13) (1000,0.13)
        };
        \addlegendentry{ \pop{}}

        \end{semilogyaxis}
    \end{tikzpicture}
    \caption{OFP Very-Hard}
    \label{fig:pg_runtime_all_methods_OFP_very_hard}
\end{subfigure}

\caption{Average PG Evolution across Methods and Problem Instances.}
\label{fig:PG_over_time}
\end{figure}

\noindent \textbf{ \pop{} Performance on Instances solved to a PG below 0.05\%.} Figure \ref{fig:PG_over_time} shows the PG over time, averaged over 100 test instances on each benchmark for each of the four methods. The figure reports the evolution of the PG (in \%) over time (0–1000 seconds) on a logarithmic scale for eight instances across five problem classes (MMCNP, SLAP, COURSE, OCP, and OFP), each in Hard and Very-Hard variants. Each subfigure compares the four methods by tracking the rate and magnitude of reduction in the PG.

Across almost all instances,  \pop{} exhibits a markedly faster initial reduction in the PG. In several cases (e.g., OCP Hard, MMCNP Hard, SLAP Hard),  \pop{} reaches sub-1\% PGs within the first 100–200 seconds and often stabilizes near 0.01–0.1\% well before 600 seconds. This early dominance indicates that  \pop{} is particularly effective at rapidly constructing high-quality feasible solutions. In contrast, MIP (Gurobi) typically reduces the PG more gradually, with a noticeable delay before reaching the same low-PG regime. \textsc{PROPEL} and \textsc{ID-PaS+} consistently outperform the vanilla MIP baseline in the intermediate time window (roughly 100–400 seconds), often reducing the PG substantially earlier and sometimes matching  \pop{} in final solution quality, though typically not in early speed.

The performance differences become more pronounced on the Very-Hard instances. For OCP Very-Hard and OFP Very-Hard, MIP and even \textsc{PROPEL} and \textsc{ID-PaS+} tend to plateau at non-negligible PGs. At the same time,  \pop{} continues to significantly reduce the PG, particularly in the mid-to-late runtime regime. A similar pattern is observed in MMCNP and SLAP instances, where \textsc{PROPEL} and \textsc{ID-PaS+} consistently improve upon MIP but are slightly less aggressive in early convergence than  \pop{}. Overall, the graphs suggest a clear hierarchy:  \pop{} dominates in early feasible-solution quality and often in final PG, \textsc{PROPEL} and \textsc{ID-PaS+} provide stable improvements over MIP, and the baseline MIP solver exhibits the slowest PG closure, particularly as instance difficulty increases.

\noindent \textbf{ \pop{} Performance on Instances solved with PG higher than 0.05\% or infeasible.} Table \ref{tab:correction} reports the decomposition of the method into its successive phases: the baseline MIP configuration, completion, and up to four Correction iterations (Phase III). For each problem class and difficulty level, the table shows the final PG after a fixed runtime budget, allowing a clear attribution of performance gains to each stage. The instances range from moderately difficult Hard cases to substantially more challenging Very-Hard cases, providing a consistent view of how the refinement mechanism behaves as complexity increases.

The completion accounts for the majority of the improvement, while corrections primarily serve as a robustness mechanism that recovers from occasional prediction errors and further refines solution quality. Across all problem classes, the PG decreases substantially from the baseline MIP to completion. For example, in OCP Very-Hard, the PG decreases from 3.39 to 0.93, and in MMCNP Very-Hard from 1.54 to 0.12. Similar patterns appear in OFP Very-Hard (1.39 to 0.21) and OCP Hard (0.51 to 0.12). This shows that most of the performance improvement is already captured by the completion, which is highly effective at identifying strong regions of the parameter space before any iterative correction is applied.

\begin{table}[!t]
  \centering
  \caption{Evolution of PG across the proposed \pop{} framework, including completion and corrections. Results are averaged over instances not solved within the desired threshold.}
    \begin{adjustbox}{width=\textwidth}
    \begin{tabular}{l|c|c|c|c|c|c|c}
    \toprule
    Problem & Instances & MIP   & Completion & Correction 1 & Correction 2 & Correction 3 & Correction 4 \\
    \midrule
    MMCNP Hard & 21    & 0.41  & 0.15  & 0.11  & 0.07  & 0.07  & 0.07 \\
    \midrule
    MMCNP Very-Hard & 19    & 1.54  & 0.12  & 0.08  & 0.06  & 0.06  & 0.06 \\
    \midrule
    SLAP Hard & -     & 0.10 & 0.03 & -     & -     & -     & - \\
    \midrule
    SLAP Very-Hard & 16    & 0.39  & 0.15  & 0.11  & 0.07  & 0.07  & 0.07 \\
    \midrule
    COURSE Hard & -     & 0.10 & 0.03 & -     & -     & -     & - \\
    \midrule
    COURSE Very-Hard & 16    & 0.39  & 0.15  & 0.11  & 0.07  & 0.07  & 0.07 \\
    \midrule
    OCP Hard & 15    & 0.51  & 0.12  & 0.07  & 0.07  & 0.07     & 0.07 \\
    \midrule
    OCP Very-Hard & 17    & 3.39  & 0.93  & 0.80  & 0.80  & 0.80  & 0.80 \\
    \midrule
    OFP Hard & 24    & 0.71  & 0.14  & 0.10  & 0.07  & 0.07  & - \\
    \midrule
    OFP Very-Hard & 26    & 1.39  & 0.21  & 0.15  & 0.10  & 0.07  & 0.07 \\
    \bottomrule
    \end{tabular}%
  \label{tab:correction}%
  \end{adjustbox}
\end{table}

The corrections then deliver systematic but diminishing improvements. The first one or two corrections typically produce the largest additional reductions, while later iterations provide finer refinements. In OFP Very-Hard, the PG decreases from 0.21 to 0.07 over three corrections. In MMCNP Hard, the PG decreases from 0.15 to 0.07 within two corrections. In easier cases, such as SLAP Hard, the completion alone already yields a very small PG (0.03), making further corrections unnecessary. Overall, the table highlights a two-stage effect: the completion delivers the dominant improvement, while the correction iterations act as a controlled polishing mechanism that incrementally consolidates solution quality, especially on the most difficult instances.

By progressively enabling the completion and correction mechanisms, supervised learning and targeted corrections jointly improve feasibility, optimality, and robustness across problem classes.

\noindent \textbf{Impact of Domain-specific Features.} Table \ref{tab:features} evaluates the impact of the feature design used in the ML model. Two feature sets are compared: $F_1$, which includes only generic instance features, and $F_2$, which augments $F_1$ with domain-specific features. For each problem class and difficulty level, the table reports the mean and standard deviation of two performance indicators: the PG and the PI. This allows assessment not only of average solution quality but also of robustness across instances.

\begin{table}[!t]
  \centering
  \caption{Impact of feature engineering on ML model performance. Comparison between the generic feature set $F_1$ and the enriched feature set $F_2$ in terms of PG and PI, reported as mean and standard deviation across instances.}
  \begin{adjustbox}{width=0.9\textwidth}
    \begin{tabular}{l|cc|cc|cc|cc}
    \hline
    \multicolumn{1}{c|}{\multirow{3}[5]{*}{Problem}} & \multicolumn{4}{c|}{$F_1$}       & \multicolumn{4}{c}{$F_2$} \\
\cmidrule{2-9}          & \multicolumn{2}{c|}{PG (\%) $\downarrow$} & \multicolumn{2}{c|}{PI  $\downarrow$} & \multicolumn{2}{c|}{PG (\%) $\downarrow$} & \multicolumn{2}{c}{PI  $\downarrow$} \\
\cmidrule{2-9}          & Mean  & Std Dev & Mean  & Std Dev & Mean  & Std Dev & Mean  & Std Dev \\
    \midrule
    MMCNP Hard & 0.07  & 0.14  & 0.61  & 1.74  & \textbf{0.05} & \textbf{0.10} & \textbf{0.51} & \textbf{1.45} \\
    \midrule
    MMCNP Very-Hard & 0.056 & 0.112 & 3.24  & 0.80  & \textbf{0.04} & \textbf{0.08} & \textbf{2.70} & \textbf{0.67} \\
    \midrule
    SLAP Hard & 0.0154 & 0.0196 & 0.50  & 0.19  & \textbf{0.011} & \textbf{0.014} & \textbf{0.42} & \textbf{0.16} \\
    \midrule
    SLAP Very-Hard & 0.112 & 0.098 & 1.49  & 0.53  & \textbf{0.08} & \textbf{0.07} & \textbf{1.24} & \textbf{0.44} \\
    \midrule
    COURSE Hard & 0.008 & 0.002  & 1.77  & 2.01  & \textbf{0.004} & \textbf{0.002} & \textbf{1.43} & \textbf{1.55} \\
    \midrule
    COURSE Very-Hard & 0.07 & 0.23 & 7.12  & 4.63  & \textbf{0.05} & \textbf{0.11} & \textbf{6.76} & \textbf{4.22} \\
    \midrule
    OCP Hard & 0.098 & 0.07  & 3.85  & 1.87  & \textbf{0.07} & \textbf{0.05} & \textbf{3.21} & \textbf{1.56} \\
    \midrule
    OCP Very-Hard & 0.88  & 0.042 & 4.55  & 1.46  & \textbf{0.80} & \textbf{0.03} & \textbf{3.79} & \textbf{1.22} \\
    \midrule
    OFP Hard & 0.126 & 0.07  & 0.95  & 0.32  & \textbf{0.09} & \textbf{0.05} & \textbf{0.79} & \textbf{0.27} \\
    \midrule
    OFP Very-Hard & 0.182 & 0.112 & 1.39  & 0.30  & \textbf{0.13} & \textbf{0.08} & \textbf{1.16} & \textbf{0.25} \\
    \bottomrule
    \end{tabular}%
  \label{tab:features}%
  \end{adjustbox}
\end{table}

Across all problem classes and difficulty levels, $F_2$ consistently outperforms $F_1$ in terms of mean PG. The improvement is particularly pronounced on Very-Hard instances. For example, in OCP Very-Hard, the mean PG decreases from 0.88 ($F_1$) to 0.80 ($F_2$), and in OFP Very-Hard from 0.182 to 0.13. Similar reductions are observed for MMCNP and SLAP variants. These results indicate that incorporating domain-specific structure enables the ML model to capture instance heterogeneity better and select more appropriate parameter configurations. The performance gains are systematic rather than isolated, as $F_2$ dominates $F_1$ in every row of the table. A Wilcoxon signed-rank test ($p < 0.01$)~\citep{woolson2007wilcoxon} confirms that all reported performance differences between domain-generic and domain-specific features in  \pop{} are statistically significant, validating the robustness of the observed improvements across the five case studies. The benefits of domain-specific features are more pronounced on the very-hard instances, suggesting that instance-specific structural information becomes increasingly valuable as combinatorial complexity grows.

\subsection{Managerial Insights}\label{sec:managerial_insights}

The computational study reveals three consistent findings across all five application domains. First, each parametric MIP family exhibits a small computational backdoor whose fixation dramatically reduces solution effort, providing empirical evidence that a limited subset of variables drives most of the optimization complexity. Second, these backdoors are highly sparse, with the vast majority of backdoor variables taking value zero at optimality and only a small fraction of non-zero variables determining the key decisions. Third, fixing only the non-zero backdoor variables recovers most of the computational benefit obtained by fixing the entire backdoor, while fixing only the zero-valued variables yields substantially weaker improvements. These results suggest that the primary challenge in large-scale parametric mixed-integer optimization is not predicting all integer variables, but accurately identifying and predicting the small subset of non-zero backdoor variables that govern solution quality and computational performance. Collectively, these findings provide the empirical foundation for the \pop{} framework and motivate its predict-complete-correct paradigm.

These computational findings translate into actionable managerial insights. This section discusses how \pop{} supports faster and more reliable decision-making, identifies critical variables driving complexity, and enables scalable planning in high-dimensional operational settings.

\noindent \textbf{Faster and more reliable decision support.}
Across all problem classes, \pop{} consistently reduces both the primal gap and the primal integral, indicating not only higher final solution quality but also substantially faster convergence to high-quality feasible plans. From a managerial perspective, this means that decision-makers can obtain strong, implementable solutions earlier within limited planning windows. In time-sensitive operational environments such as port operations, production planning, or logistics coordination, earlier access to high-quality plans directly translates into improved responsiveness, reduced operational risk, and greater confidence in execution.

\noindent \textbf{Focusing on what truly drives complexity.}
A central insight of \pop{} is that not all decision variables contribute equally to computational difficulty. Rather than attempting to simplify the entire model uniformly, \pop{} concentrates on
a backdoor, i.e., a subset of structurally critical variables that drive combinatorial complexity. By accurately predicting both active and inactive values for backdoor variables, \pop{} reshapes the search space around its most influential decisions. Managerially, this provides a clear message: complexity in large-scale planning problems is often concentrated in a small number of pivotal choices, such as assignment, sequencing, or allocation decisions. Identifying and structuring these key drivers enables disproportionate gains in solvability and performance.

\noindent \textbf{Scalable planning through feature-driven tuning.}
The effectiveness of \pop{} relies on interpretable, domain-specific features that summarize instance characteristics such as demand intensity, resource capacity, structural size, and seasonality. These features serve as a compact representation of the operational environment and enable the model to adapt its parameter selection to the instance's difficulty. Importantly, the framework is modular: if additional domain knowledge becomes available, new features can be incorporated without redesigning the optimization model itself. This flexibility enables organizations to refine performance as more operational data is collected over time.

\noindent \textbf{The Elephant and the Bee analogy.}
Conceptually, \pop{} embodies a divide-and-focus philosophy. The large-scale optimization model is like an elephant: powerful, with generic features, but computationally intensive. The learning component with the domain-specific features acts as the bee: lightweight, adaptive, and capable of quickly identifying where attention should be directed. By guiding the solver toward structurally critical decisions rather than uniformly reducing the model, the approach balances robustness with efficiency. This synergy enables organizations to maintain rich, high-fidelity models while achieving tractable solution times, even as problem size and dimensionality increase.

Overall, the results suggest that data-informed variable prioritization can substantially enhance the scalability of advanced planning models. For managers, this means that complex, high-dimensional decision problems need not be simplified at the modeling stage to remain solvable. Instead, intelligent guidance layered on top of detailed optimization models can deliver both realism and computational performance.

\section{Conclusion}\label{sec:conclusion}

This paper introduced the \pop{} framework for efficiently solving parametric mixed-integer optimization problems by identifying backdoor variables that drive computational complexity and exploiting this structure through targeted prediction and correction. The framework consists of three phases: backdoor identification, which analyzes the impact of variable fixing to identify a backdoor; supervised learning, which predicts values or narrower domains for backdoor variables to produce a reduced problem that can be solved efficiently; and completion and correction, which restore feasibility and improve solution quality when predictions are imperfect. To guide the design of Phases II and III, a taxonomy was proposed based on problem size, backdoor variable type, and feature type, providing a systematic map for selecting appropriate modeling and correction strategies. The methodology leverages this taxonomy to handle binary and general integer variables, sparse and dense structures, and domain-specific features, thereby balancing computational efficiency with solution quality.

Computational experiments on five large-scale applications demonstrated that the proposed framework consistently identifies meaningful and stable backdoors corresponding to the primary operational decisions of each problem. The results show that combining prediction, completion, and correction can significantly improve solution quality relative to both classical optimization and existing learning-augmented optimization approaches. In particular, the experiments highlight that much of the computational complexity of large-scale parametric MIPs is concentrated in a small subset of strategic variables, and that accurately predicting these variables can dramatically reduce the effective search space while preserving high-quality solutions.

More broadly, the results suggest that learning and optimization are particularly complementary in repetitive decision-making environments where large numbers of related instances must be solved over time. By leveraging historical solutions to identify and predict the most influential decisions, \pop{} transforms optimization from solving each instance independently to exploiting recurring structural patterns across instances. This perspective provides a scalable pathway for integrating machine learning into mathematical optimization and establishes a foundation for future research on adaptive backdoor discovery, richer prediction models, and learning-enhanced optimization frameworks for large-scale industrial applications. The present work assumes that instances share a common structural formulation and differ primarily through parameter perturbations. Extending \pop{} to variable-size instances, where nodes, arcs, products, or constraints may be added or removed over time, constitutes an important direction for future research. Such extensions would require learning representations that generalize across varying optimization structures and could further broaden the applicability of learning-enhanced optimization in industrial settings.

\section*{Acknowledgments}

This research was partly supported by the NSF AI Institute for Advances in Optimization (Award 2112533).

\begin{spacing}{1}
\typeout{}
\bibliographystyle{apalike}
\bibliography{References.bib}
\end{spacing}    

\begin{appendices}
\renewcommand{\thesection}{\Alph{section}}
\section{Backdoor Identification} 
\label{A}

\begin{table}[t!]
\centering
\caption{Phase I backdoor identification results for the Hard setting. For each application, candidate variable families were fixed to their values in a reference optimal solution, and the resulting reduced MIP was solved. Following the backdoor identification procedure of \pop{}, the variable family yielding the shortest runtime was selected as the backdoor.}
\label{tab:backdoor_identification}
\begin{adjustbox}{width=\textwidth}
\begin{tabular}{llrrrc}
\toprule
\textbf{Problem} & \textbf{Variable Family} & \textbf{\# Var} & \textbf{Runtime (s)} & \textbf{Gap (\%)} & \textbf{Selected} \\
\midrule

\multirow{2}{*}{MMCNDP}
& $x$ (flow variables) & 2,183 & \textbf{0.001} & 0.000 & \checkmark \\
& $y$ (facility variables) & 357 & 0.004 & 0.163 & \\
\midrule

\multirow{3}{*}{SLAP}
& $x$ (assignment variables) & 39,169 & \textbf{0.020} & 0.000 & \checkmark \\
& $u$ (inventory variables) & 38,769 & 0.039 & 0.466 & \\
& $y$ (fleet variables) & 8 & 50.744 & 0.410 & \\
\midrule

\multirow{8}{*}{COURSE}
& $x$ (course assignment variables) & 20,736 & \textbf{0.031} & 0.000 & \checkmark \\
& $y$ (scheduling variables) & 1,728 & 0.050 & 0.000 & \\
& $z$ (auxiliary variables) & 20,736 & 1.477 & 0.000 & \\
& $w$ (block variables) & 300 & 0.061 & 0.000 & \\
& $v$ (slot variables) & 12 & 0.064 & 0.000 & \\
& $o$ (three variables) & 1 & 1.671 & 0.000 & \\
& $r$ (B2B variables) & 2 & 9.581 & 0.000 & \\
& $l$ (triple variables) & 2 & 100.021 & 3.681 & \\
\midrule

\multirow{3}{*}{OCP}
& $x$ (vessel assignment) & 9,690 & \textbf{6.609} & 0.213 & \checkmark \\
& $y$ (order fulfillment) & 61 & 100.044 & -- & \\
& $v$ (selected variables) & 1,404 & 16.416 & 0.034 & \\
\midrule

\multirow{3}{*}{OFP}
& $x$ (allocation variables) & 288,000 & \textbf{0.131} & 0.000 & \checkmark \\
& $s$ (setup variables) & 480 & 0.962 & 0.216 & \\
& $a$ (activation variables) & 100 & 1.820 & 0.360 & \\
\bottomrule
\end{tabular}
\end{adjustbox}
\end{table}

\begin{table}[t!]
\centering
\caption{Phase I backdoor identification results for the Very-Hard setting. For each application, candidate variable families were fixed to their values in a reference optimal solution, and the resulting reduced MIP was solved. Following the backdoor identification procedure of \pop{}, the variable family yielding the shortest runtime was selected as the backdoor.}
\label{tab:backdoor_identification_vhard}
\begin{adjustbox}{width=\textwidth}
\begin{tabular}{llrrrc}
\toprule
\textbf{Problem} & \textbf{Variable Family} & \textbf{\# Var} & \textbf{Runtime (s)} & \textbf{Gap (\%)} & \textbf{Selected} \\
\midrule

\multirow{2}{*}{MMCNDP}
& $x$ (flow variables) & 23,984 & \textbf{0.010} & 0.000 & \checkmark \\
& $y$ (facility variables) & 1,135 & 100.010 & -- & \\
\midrule

\multirow{3}{*}{SLAP}
& $x$ (assignment variables) & 39,916 & \textbf{0.022} & 0.000 & \checkmark \\
& $u$ (inventory variables) & 37,116 & 0.094 & 0.000 & \\
& $y$ (fleet variables) & 20 & 100.037 & 0.537 & \\
\midrule

\multirow{8}{*}{COURSE}
& $x$ (course assignment variables) & 38,416 & \textbf{0.050} & 0.000 & \checkmark \\
& $y$ (scheduling variables) & 2,744 & 0.092 & 0.000 & \\
& $z$ (auxiliary variables) & 38,416 & 3.723 & 0.499 & \\
& $w$ (block variables) & 406 & 0.116 & 0.000 & \\
& $v$ (slot variables) & 14 & 0.121 & 0.000 & \\
& $o$ (three variables) & 1 & 4.034 & 0.000 & \\
& $r$ (B2B variables) & 2 & 21.685 & 0.000 & \\
& $l$ (triple variables) & 2 & 100.017 & 6.455 & \\
\midrule

\multirow{5}{*}{OCP}
& $x$ (vessel assignment) & 5,682 & \textbf{20.725} & 0.000 & \checkmark \\
& $y$ (order fulfillment) & 38 & 55.202 & 0.000 & \\
& $v$ (selected variables) & 1,179 & 92.761 & 0.000 & \\
& $w$ (production variables) & 4,154 & 60.009 & 0.000 & \\
& $o$ (changeover variables) & 5,239 & 100.021 & 16.161 & \\
\midrule

\multirow{3}{*}{OFP}
& $x$ (allocation variables) & 2,000,000 & \textbf{1.153} & 0.000 & \checkmark \\
& $a$ (activation variables) & 200 & 11.706 & 0.454 & \\
& $s$ (setup variables) & 1,000 & 64.206 & 0.487 & \\
\bottomrule
\end{tabular}
\end{adjustbox}
\end{table}

Tables~\ref{tab:backdoor_identification} and \ref{tab:backdoor_identification_vhard} report the Phase I exploratory analysis used to identify candidate backdoors for the hard and very-hard settings, respectively. For each application, several semantically meaningful variable families were fixed to their values in a reference optimal solution, and the resulting reduced MIP was solved. Consistent with the backdoor identification procedure described in Section~\ref{sec:framework}, the family yielding the shortest runtime was selected as the backdoor. Interestingly, the selected backdoors correspond to the principal operational decision variables of each application (transportation flows in MMCNDP, locomotive assignments in SLAP, course assignments in COURSE, vessel assignment in OCP, and order-allocation variables in OFP), providing empirical evidence that a relatively small subset of strategic decisions drives most of the computational complexity. Notably, the selected backdoor variables are also those with the largest cardinality in four of the five applications, indicating that the most expressive operational decisions tend to capture the majority of the combinatorial structure of the problem.

The results reveal substantial differences in the computational impact of the candidate variable families. In all five applications, the selected backdoor yields the smallest solution time, often by several orders of magnitude. These findings suggest that the computational complexity of the considered parametric MIPs is concentrated in a relatively small subset of strategic decision variables, which motivates the prediction, completion, and correction mechanisms developed in Phases II and III.

The results indicate that the same variable family is consistently identified as the backdoor across both Hard and Very-Hard instances for all five applications. This empirical observation provides strong support for the assumption made in Section~\ref{sec:framework} that the backdoor remains essentially stable across instances drawn from the same distribution. Moreover, the selected backdoors consistently correspond to the primary operational decision variables of each application, suggesting that the computational complexity of these parametric MIPs is concentrated in a small set of strategic decisions.

Although the selected backdoor variables often correspond to the largest variable family in the formulation, the exploratory analysis shows that cardinality alone is insufficient to explain computational impact, as several large variable families produce significantly weaker reductions in solution time.

\end{appendices}

\end{document}